%% file: paper.tex
\PassOptionsToPackage{table}{xcolor}
\documentclass[11pt]{article}
\usepackage[preprint]{acl}
\usepackage{times,latexsym}
\usepackage[T1]{fontenc}
\usepackage[utf8]{inputenc}
\usepackage{microtype,amsmath,amssymb,booktabs,graphicx,multirow,array}
\usepackage{xcolor}
\definecolor{TableHeader}{HTML}{E4EBF7}
\definecolor{TableHeaderDark}{HTML}{172B4D}
\definecolor{PCTTint}{HTML}{EAF1FF}
\definecolor{PCTTeal}{HTML}{2563EB}
\newcommand{\pct}{\textsc{HexaBee-4B}}
\newcommand{\Lset}{\mathcal{L}}
\newcommand{\Dset}{\mathcal{D}}
\title{HexaBee-4B: Six-Language Translation with\\Multilingual Answering and Bidirectional Supervision}
\author{Iman Barati$^{1}$\thanks{Corresponding author.} \quad
Mohammadmostafa Rostamkhani$^{1}$ \quad Mohammad Badzohreh$^{2}$\\
{\normalfont\small $^{1}$Iran University of Science and Technology, Tehran, Iran}\\
{\normalfont\small $^{2}$Tarbiat Modares University, Tehran, Iran}\\
{\normalfont\small\texttt{iman\_barati@comp.iust.ac.ir} \quad
\texttt{mo\_rostamkhani97@comp.iust.ac.ir}}\\
{\normalfont\small\texttt{mohammad.badzohreh@modares.ac.ir}}}
\begin{document}
\maketitle
\begin{abstract}
Multilingual translation requires a model to preserve meaning while producing text in a requested language. We present \pct{}, a translator adapted from Gemma 3 4B IT through three stages: Wikipedia continued pretraining, joint multilingual answering and answer-text translation, and bidirectional translation specialization. The intermediate stage connects two tasks through shared multilingual answers: answering a question in the requested language and translating an answer into that language. It comprises 19.46 million answering records, including 5.76 million cross-language question--answer pairs, and 10.70 million answer-text translations. The final stage uses 658,274 forward and reverse synthetic translation examples. On 36,572 synthetic test examples spanning all 30 directions among Arabic, Chinese, English, Hebrew, Persian, and Spanish, \pct{} achieves direction-macro chrF++ of 73.75, compared with 57.11 for Gemma 3 4B IT and 66.83 for Gemma 3 12B IT. BLEU, TER, COMET, and a matched model-judged subset support the same overall ranking. These results demonstrate improved translation across the six-language set, with chrF++ gains in every evaluated direction over both baselines.
\end{abstract}

\section{Introduction}
A multilingual language model can encounter the same information in several languages, yet translation requires more than recognizing that information: its response must preserve the source meaning and follow a specified target language. Multilingual neural translation establishes target-language conditioning as a central design choice \citep{johnson-etal-2017-googles}. More recent work shows that continued training and translation-oriented instruction tuning can adapt general language models into stronger translators \citep{xu2024a,li-etal-2024-eliciting}. For a compact instruction-tuned model, the organization of this supervision is therefore an important part of the translation system.

Multilingual instruction tuning supplies a broader source of supervision. Multitask training and multilingual instruction collections improve instruction following across languages \citep{muennighoff-etal-2023-crosslingual,singh-etal-2024-aya,ustun-etal-2024-aya}; even limited multilingual instruction data can support cross-language generalization \citep{shaham-etal-2024-multilingual}. However, multilingual data need not require communication \emph{between} languages. A question and its answer may always use the same language. Explicitly separating their language assignments creates a different task: understand a question in one language and answer it in another. Cross-language instruction tuning and translation-following demonstrations provide precedents for this design \citep{lin-etal-2025-crossin,ranaldi-etal-2024-empowering,zhu-etal-2024-question}.

\begin{figure*}[t]
\centering\includegraphics[width=\textwidth]{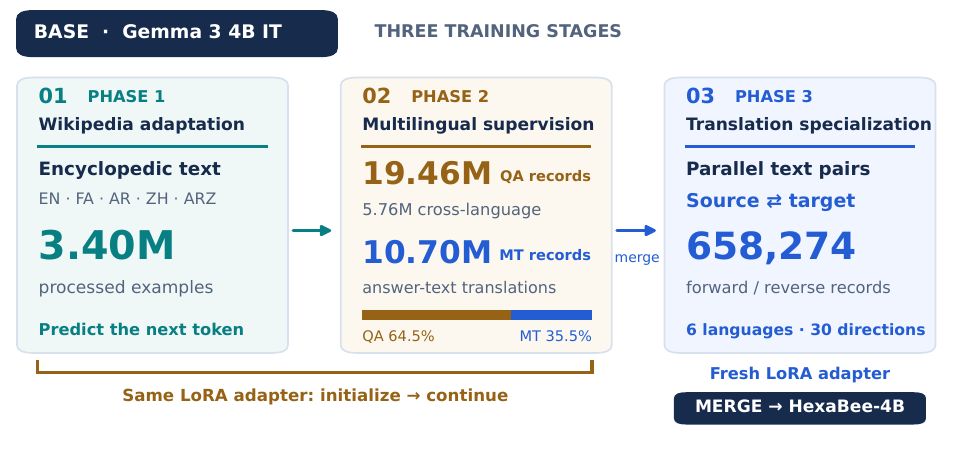}
\caption{The complete adaptation recipe. Phase 1 applies next-token language modeling to Wikipedia text. Phase 2 continues the same adapter with multilingual answering and answer-text translation. After merging, phase 3 trains a fresh adapter on bidirectional translations and produces the final merged \pct{} model. The base is Gemma 3 4B IT. Phase 1 counts processed training examples; phases 2--3 count SFT records. Examples derived from one source can occur in multiple tasks.}
\label{fig:method}\end{figure*}

\begin{figure}[t]
\centering\includegraphics[width=\linewidth]{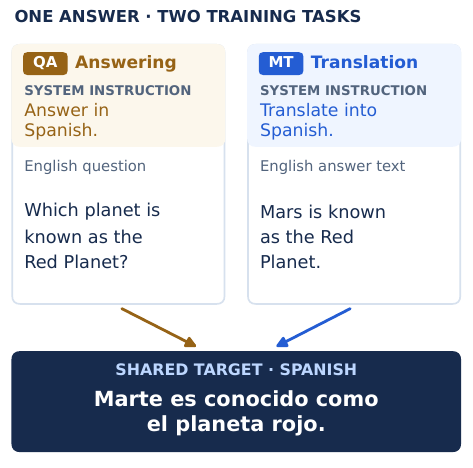}
\caption{Illustrative English-to-Spanish supervision. The tasks have different inputs and instructions but can share the same target answer. This constructed example explains the method; it is not a quoted training record.}
\label{fig:sharedanswer}\end{figure}

We investigate how to organize such answering supervision within a translation-focused adaptation recipe (Figure~\ref{fig:method}). Consider an English question whose requested answer is Spanish. The model must answer the question in Spanish, whereas an English-to-Spanish translation task must preserve the content of its English input. Our construction connects these tasks through multilingual answer texts: a Spanish answer can supervise both an English question and an English version of that answer. Figure~\ref{fig:sharedanswer} illustrates this shared-target relation. The system instruction specifies the desired response language and, where relevant, response style. Its own wording can vary across languages, but this variation is a supporting data-construction choice. Translation remains the primary task evaluated here.

This separation of language roles is also motivated by multilingual question-answering research. TyDi QA and MKQA broaden the linguistic coverage of QA evaluation \citep{clark-etal-2020-tydi,longpre-etal-2021-mkqa}. XHotpotQA makes language assignments to questions and evidence roles explicit when studying cross-lingual knowledge composition \citep{barati2026xhotpotqa}. Our work applies the related principle of separating language roles to training question--answer generation and answer-text translation. The resulting supervision uses a Wikipedia-derived multilingual inventory and does not require retrieving evidence at inference time.

We introduce \pct{}, built from \textbf{Gemma 3 4B IT} \citep{gemmateam2025gemma3}.\footnote{Resources: \href{https://github.com/Iman998/HexaBee-4B}{Code}; \href{https://huggingface.co/Iman998/HexaBee-4B}{Final model}.} Its three stages provide encyclopedic language modeling, multilingual response supervision, and translation specialization. The second stage combines 19,458,332 answering records with 10,695,072 answer-text translations; 5,759,806 answering records cross question and response languages. The third stage uses 658,274 bidirectional synthetic examples. We evaluate the complete model across six languages using five automatic metrics and paired model-based judgments. Our contributions are the three-stage translation recipe, an explicit account of how answering and translation share supervision, and a direction-complete evaluation against public Gemma 3 4B IT and 12B IT. The comparison measures the complete recipe; it does not isolate the effect of an individual training component.

\section{Related Work}
\label{sec:related}
\paragraph{Multilingual translation and adaptation.}
Multilingual MT shares parameters across translation directions and uses explicit target-language signals \citep{johnson-etal-2017-googles}. ALMA studies translation-oriented continued training of large language models \citep{xu2024a}, while \citet{li-etal-2024-eliciting} examine multilingual finetuning with translation instructions. Our staged design combines this translation focus with an intermediate answering task. Its initial Wikipedia adaptation follows the broader motivation of continued pretraining on relevant text \citep{gururangan-etal-2020-dont}. Training on teacher-generated translations is related to sequence-level knowledge distillation \citep{kim-rush-2016-sequence}.

\paragraph{Multilingual instruction supervision.}
Multitask finetuning enables cross-lingual generalization \citep{muennighoff-etal-2023-crosslingual}, and Aya provides multilingual instruction data and models \citep{singh-etal-2024-aya,ustun-etal-2024-aya}. CrossIn combines cross-language instruction--response pairs with instruction translation \citep{lin-etal-2025-crossin}. CrossAlpaca uses instruction-following and translation-following demonstrations \citep{ranaldi-etal-2024-empowering}; Question Translation Training aligns reasoning questions across languages \citep{zhu-etal-2024-question}. These studies establish relevant precedents for coupling instruction following and translation. Our contribution is the particular combination of system-directed answering, translations of answer texts, and subsequent bidirectional specialization on Gemma 3 4B IT, together with its measured corpus composition and translation results.

\paragraph{Language roles in question answering.}
TyDi QA collects information-seeking questions directly in diverse languages \citep{clark-etal-2020-tydi}, while MKQA supports comparison across languages through aligned questions and language-independent answer representations \citep{longpre-etal-2021-mkqa}. XHotpotQA studies mixed-language evidence in multi-hop QA by assigning languages to distinct roles in an evidence-dependency graph \citep{barati2026xhotpotqa}. This provides a conceptual motivation for making language boundaries explicit rather than treating multilinguality only as a collection of monolingual examples. Here, the relevant roles are the question, requested answer, and instruction wording; we use their combinations as generation supervision for a translator. Our experiments concern translation quality, whereas those benchmarks evaluate answering or knowledge composition.

\section{Method}
\label{sec:task}
\subsection{Answering and translation as connected tasks}
Let $\Lset=\{\mathrm{ar},\mathrm{zh},\mathrm{en},\mathrm{he},\mathrm{fa},\mathrm{es}\}$ be the six translation languages. We use $s,t\in\Lset$ for input and response languages. For a content item $c$, let $q_s(c)$ denote its question in language $s$, and let $a_t^v(c)$ denote an answer in language $t$ with response variant $v$ (extent or style). Let $S$ denote the \emph{system instruction text}. Specifically, $S^{\mathrm{QA}}_{t,v}$ requests an answer in language $t$ with variant $v$, and $S^{\mathrm{MT}}_{s,t}$ requests translation from $s$ into $t$.

The two supervision relations are
\begin{align}
\mathrm{QA}:\quad &(S^{\mathrm{QA}}_{t,v},q_s(c))\mapsto a_t^v(c),\label{eq:qa}\\
\mathrm{MT}:\quad &(S^{\mathrm{MT}}_{s,t},a_s(c))\mapsto a_t(c),\label{eq:mt}
\end{align}
where $a_s(c)$ and $a_t(c)$ are the answer variants selected for translation. QA includes same-language and cross-language pairs; MT uses different input and output languages. A shared answer can therefore appear as the target of both tasks. Its input is a question in QA and an answer text in MT. This relation connects response generation to semantic transfer without adding an intermediate generation step.

Instruction pools provide alternative phrasings for response language, extent, style, and translation direction.\footnote{Resources: \href{https://huggingface.co/datasets/Iman998/HexaBee-System-Prompts}{System prompts}.} Cross-language QA explicitly requests the answer language; some same-language examples request only answer extent. A directive's wording language and its requested response language are separate choices. We preserve this distinction when constructing the data, rather than assuming that a prompt written in one language always requests that language. Pool composition and sampling are documented in Appendix~\ref{app:weights}.

\subsection{A shared response objective}
For either task, let $x$ be the input text and $y=(y_1,\ldots,y_m)$ the target response of $m$ tokens. With model parameters $\theta$, the response-token negative log-likelihood is
\begin{equation}
\ell_\theta(S,x,y)=-\sum_{k=1}^{m}\log P_\theta(y_k\mid S,x,y_{<k}),
\label{eq:conditional}
\end{equation}
where $y_{<k}$ is the response prefix preceding token $k$. The instruction and input are conditioning text; the response supplies supervision. QA and translation examples use this same objective with their respective inputs and targets. Let $\Dset_{\mathrm{QA}}$ and $\Dset_{\mathrm{MT}}$ denote the retained QA and translation records. The phase-2 dataset is the multiset union $\Dset_2=\Dset_{\mathrm{QA}}\uplus\Dset_{\mathrm{MT}}$, so repeated or reused examples retain their contribution to the training mixture. We do not introduce a separate alignment loss between the tasks.

\section{Three-Stage Training Data and Adaptation}
\label{sec:data}\label{sec:training}
\subsection{Phase 1: Wikipedia continued pretraining}
The first stage adapts Gemma 3 4B IT to encyclopedic text before introducing the paired generation tasks. It uses Wikipedia text in English, Persian, Arabic, Chinese, and Egyptian Arabic. These are the languages registered for this stage; the later six-language translation set additionally includes Hebrew and Spanish. The corpus is monolingual text, so this phase provides language-modeling supervision rather than question--answer or translation pairs.

For a tokenized sequence $w=(w_1,\ldots,w_n)$, its autoregressive objective is
\begin{equation}
\ell^{\mathrm{PT}}_\theta(w)=-\sum_{k=2}^{n}\log P_\theta(w_k\mid w_{<k}),
\label{eq:pt}
\end{equation}
where $n$ is sequence length and $w_{<k}$ contains preceding tokens. In contrast to Equation~\ref{eq:conditional}, supervision is applied throughout the text sequence, without a separate requested response. This stage aims to adapt vocabulary use and encyclopedic text modeling to the material used in the next stage, consistent with the motivation of continued pretraining \citep{gururangan-etal-2020-dont}.

A LoRA adapter \citep{hu2022lora} is trained for one epoch on 3,399,664 processed training examples. The available raw source files contain 11,395,061 text rows before preprocessing; these counts have different units. We retain the trained adapter for phase 2.\footnote{Resources: \href{https://huggingface.co/datasets/Iman998/HexaBee-Wikipedia-PT}{Wikipedia corpus}; \href{https://huggingface.co/Iman998/HexaBee-4B-PT-LoRA}{Phase-1 adapter}.} Appendix~\ref{app:ptdata} gives source-language counts and the repeated-source composition of the original mixture.

\subsection{Phase 2: multilingual answering and answer translation}

\begin{figure*}[t]
\centering\includegraphics[width=\textwidth]{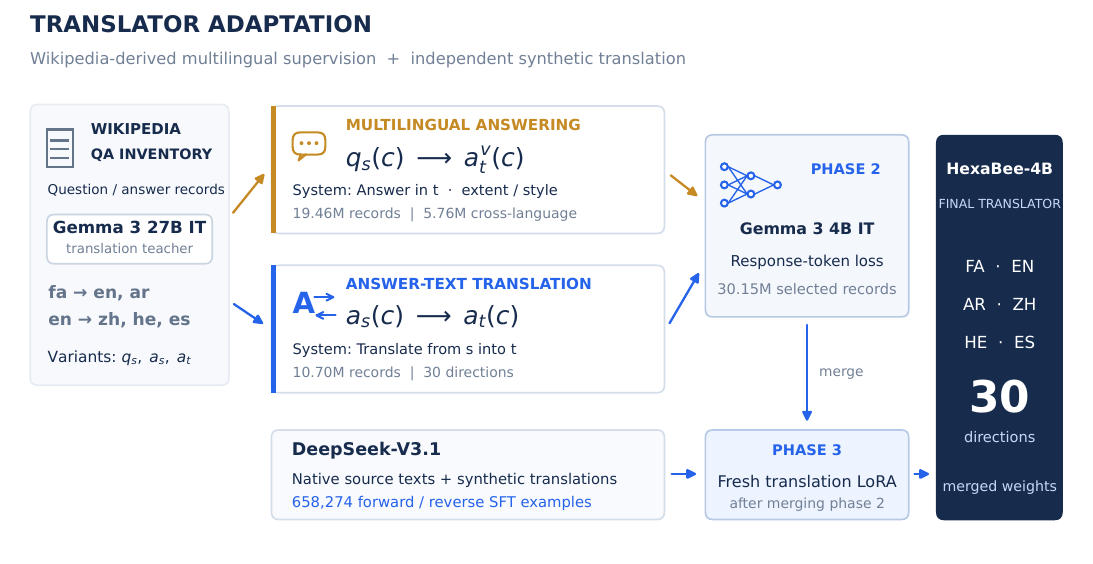}
\caption{Data and supervision flow in phases 2 and 3. Gemma 3 27B IT expands the Wikipedia-derived inventory into multilingual question and answer variants. Answering and answer-text translation supply complementary phase-2 inputs, with system instructions specifying the requested response. A separate DeepSeek-generated corpus supplies the final bidirectional translation stage. The full checkpoint sequence, including phase 1, is shown in Figure~\ref{fig:method}.}
\label{fig:supervisionflow}\end{figure*}

The second stage constructs multilingual question and answer variants from 2,025,730 Wikipedia-derived source rows. The source inventory begins with Persian content and is expanded into the other languages using Gemma 3 27B IT \citep{gemmateam2025gemma3}. Figure~\ref{fig:supervisionflow} traces this inventory into the two supervision branches. The resulting multilingual inventory, rather than a single language, defines the supervision in Equations~\ref{eq:qa}--\ref{eq:mt}. Construction provenance and translation paths are given in Appendix~\ref{app:construction}.

For answering, the input is a question and the target is an answer selected in the requested language. The variants include complete and moderate answers, as well as formal and informal responses. For answer-text translation, both sides are answer variants associated with the same content. All constructed answer-translation groups participate in this phase alongside QA; translation is therefore a substantial part of the intermediate training, not only the final specialization.

\input{results/phase2_main_table.tex}

The retained mixture contains 30,153,404 records\footnote{Resources: \href{https://huggingface.co/datasets/Iman998/HexaBee-Multilingual-SFT}{Phase-2 training data}.}: 19,458,332 answering records and 10,695,072 answer-text translations (Table~\ref{tab:phase2}). Cross-language answering accounts for 5,759,806 records, or 29.6\% of QA supervision. Answering and translation account for 64.5\% and 35.5\% of the full mixture, respectively. These are corpus proportions, not additional loss weights. The same source item can contribute to several tasks or language combinations.

The QA distribution is asymmetric: the complete-answer family covers 27 cross-language directions, while the combined translation data cover all 30. Retained data also contain empty fields and a small set of conflicting language directives. We document their counts and affected directions in Appendix~\ref{app:phase2}. The intended task is language-controlled answering; the recorded corpus is an imperfect realization of that design.

Phase 2 continues the phase-1 adapter for one epoch on the combined answering/translation mixture. The resulting adapter is then merged with Gemma 3 4B IT to obtain the multilingual checkpoint used by the final stage.\footnote{Resources: \href{https://huggingface.co/Iman998/HexaBee-4B-Multilingual-SFT-LoRA}{Phase-2 adapter}; \href{https://huggingface.co/Iman998/HexaBee-4B-Multilingual-Base}{Merged multilingual checkpoint}.} This continuity allows both phases to contribute to the checkpoint before translation specialization.

\subsection{Phase 3: bidirectional translation specialization}
The third stage uses a separately generated translation corpus. DeepSeek-V3.1 source-generation metadata \citep{deepseekai2025v31,deepseekai2024v3} identifies the teacher used to produce texts across topics, genres, registers, and complexity levels in the six languages. Each source text is paired with a synthetic translation into a different language. Native source generation broadens the content beyond the Wikipedia-derived answering inventory; its diversity is a construction goal, not evidence of coverage of every deployment domain.

The corpus contains 365,718 aligned pairs, with 60,953 source texts per language. Its stored split contains 329,146 training pairs and 36,572 test pairs. Filtering leaves 329,137 training pairs for bidirectional supervision. Each retained pair supplies two examples: translating the source into its paired target and translating the paired target back into the original source language. Thus the final SFT dataset contains $2\times329{,}137=658{,}274$ records.\footnote{Resources: \href{https://huggingface.co/datasets/Iman998/HexaBee-Bidirectional-Translation}{Bidirectional translation corpus}.} Both orientations share an underlying pair. The reverse example reuses the existing texts; it does not require another teacher call.

The use of decoded teacher translations is related to sequence-level distillation \citep{kim-rush-2016-sequence}. Here, both source generation and target translation are synthetic, so evaluation on the held-out portion measures this particular distribution. The exact split and filtering details are retained in Appendix~\ref{app:construction}.

A fresh LoRA adapter is trained on the merged phase-2 checkpoint using the response objective in Equation~\ref{eq:conditional}. After one epoch, it is merged to produce \pct{}.\footnote{Resources: \href{https://huggingface.co/Iman998/HexaBee-4B-Translation-LoRA}{Phase-3 adapter}.} All three stages use LLaMA-Factory \citep{zheng-etal-2024-llamafactory}; adapter parameterization, optimization settings, and serialization are in Appendix~\ref{app:training}. The complete sequence is therefore continued pretraining, joint answering/translation SFT, and translation-only SFT, with two adapter merges.

\section{Experimental Setup}
\label{sec:eval}
\subsection{Evaluation data and comparison systems}
We evaluate stored outputs for all 36,572 held-out synthetic examples, covering the 30 directed language pairs. Each direction contains 1,173--1,276 examples. The comparison includes \pct{}, public Gemma 3 4B IT, and public Gemma 3 12B IT \citep{gemmateam2025gemma3}; all three use the same source texts and references. A matched subset of 300 examples, ten per direction, is additionally scored by a model judge.

The archived inference templates differ between \pct{} and the public baselines. Consequently, results compare the recorded systems, including their prompts, rather than isolating adaptation under an identical prompt. We score outputs as stored, including empty responses and unsolicited text. Inference settings and failure counts are reported in Appendix~\ref{app:inference}. No human-authored translation benchmark is evaluated in this study; FLORES provides a complementary evaluation setting for future validation \citep{goyal-etal-2022-flores}.

\subsection{Automatic metrics}
We report BLEU \citep{papineni-etal-2002-bleu}, chrF \citep{popovic-2015-chrf}, chrF++ \citep{popovic-2017-chrf}, TER \citep{snover-etal-2006-study}, and COMET \citep{rei-etal-2020-comet}. SacreBLEU provides reproducible lexical-metric settings \citep{post-2018-call}; COMET uses the fixed \texttt{wmt22-comet-da} model \citep{unbabel2022cometda}.\footnote{Resources: \href{https://huggingface.co/datasets/Iman998/HexaBee-Model-Outputs}{Stored model outputs}; \href{https://huggingface.co/datasets/Iman998/HexaBee-Evaluation}{Evaluation tables and figures}.} TER is lower-is-better; the other reported metrics are higher-is-better. COMET values are multiplied by 100 for display.

Let $\Omega=\{(s,t)\in\Lset^2:s\ne t\}$ be the set of 30 directions and $M_d$ the score of a metric computed on direction $d$. Our primary summary is
\begin{equation}
M_{\mathrm{macro}}=\frac{1}{|\Omega|}\sum_{d\in\Omega}M_d.
\label{eq:macro}
\end{equation}
This gives every direction equal weight. A target-language summary averages its five incoming directions. Pooled corpus scores are different aggregations and are reported separately in Appendix~\ref{app:metrics}, together with metric settings, encoder truncation, and every directional result.

\subsection{Model judging and uncertainty}
Model-based translation evaluation can complement reference-overlap metrics \citep{kocmi-federmann-2023-large,kocmi-federmann-2023-gemba}. We use the archived DeepSeek judge in two modes: reference-assisted, with the synthetic reference, and reference-free, with source and candidate only. Both modes produce a holistic score from 0 to 100 and eight rubric dimensions covering accuracy, terminology, fluency, cohesion, context, language adherence, style, and formatting. These are model judgments, not calibrated accuracy percentages or human annotations.

We compare systems on the same 300 examples and estimate 95\% intervals for score differences using paired bootstrap resampling within each translation direction \citep{koehn-2004-statistical}. This preserves the matched comparison and gives each direction the same weight. The intervals capture item-sampling uncertainty on this subset; they do not quantify variability across training runs or alternative judges. Appendix~\ref{app:uncertainty} gives the procedure and rubric details.

\section{Results and Analysis}
\label{sec:results}
\input{results/main_comparison.tex}
\subsection{Overall translation quality}
\pct{} obtains the best overall score on all five automatic metrics (Table~\ref{tab:main}). Direction-macro chrF++ reaches 73.75, compared with 57.11 for Gemma 3 4B IT and 66.83 for Gemma 3 12B IT. This is a gain of 16.64 points over the same-size public base and 6.92 points over the larger comparator. BLEU increases from 40.70 to 62.03 relative to the 4B baseline; TER decreases from 51.40 to 30.05, and COMET increases from 85.88 to 89.86. The agreement between lexical and learned metrics supports improved translation on the evaluated synthetic distribution.

\begin{figure*}[t]
\centering\includegraphics[width=.95\textwidth]{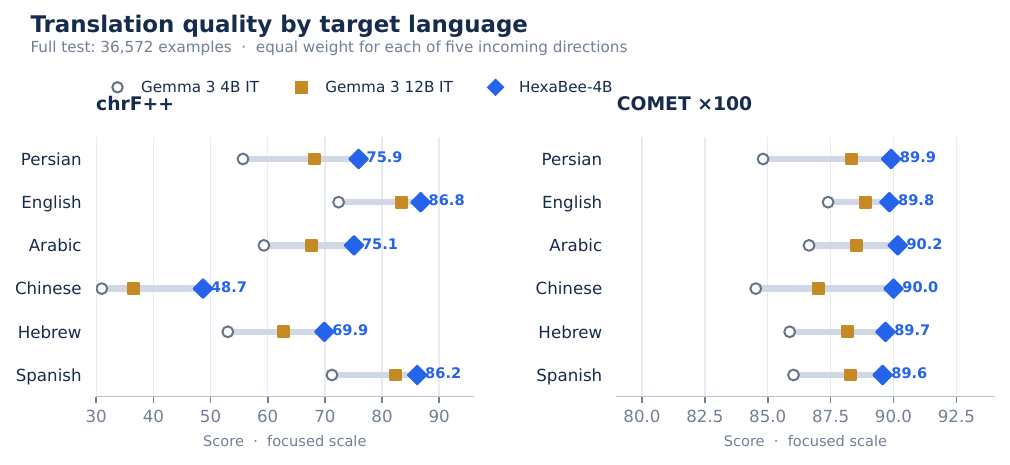}
\caption{Translation quality by target language. Each point averages five incoming directions on the full test partition. \pct{} improves both direction-macro chrF++ and COMET for each target language. The focused axis ranges are shown explicitly; marker shapes distinguish systems as well as colors.}
\label{fig:targets}\end{figure*}

\subsection{Language and direction differences}
The chrF++ gain is positive in every direction relative to both baselines. Figure~\ref{fig:targets} shows improvements for each target language, while Appendix~\ref{app:metrics} reports all source--target comparisons. Relative to the 4B baseline, the largest directional gain is Chinese-to-Persian (+27.09), and the smallest is English-to-Spanish (+6.64). Thus improvement extends beyond one target language or one script pairing.

Absolute chrF++ remains lower for Chinese-target translations than for the other target groups. This should not be read as a calibrated ranking of language difficulty: script, segmentation, reference phrasing, and the metric's character/word components affect cross-language score comparisons. The metric agreement is also not universal at the direction level. For Chinese-to-Hebrew, \pct{} has TER 90.09, worse than 88.40 for 4B IT and 78.13 for 12B IT, despite higher BLEU and chrF++. TER improves in the other 29 directions. This exception motivates inspecting multiple metrics rather than inferring uniform improvements from a single aggregate.

\subsection{Judged quality}
Both judge modes follow the automatic-metric ranking. Reference-assisted means are 92.92 for \pct{}, 80.32 for 4B IT, and 91.73 for 12B IT; reference-free means are 94.10, 81.43, and 91.97. Table~\ref{tab:ci} reports paired differences and uncertainty. The comparison with the larger model is considerably closer than the comparison with the 4B base.

\begin{table}[t]
\centering\small
\begin{tabular}{llrr}
\toprule
\rowcolor{TableHeaderDark} {\color{white}\bfseries Baseline} & {\color{white}\bfseries Judge view} & {\color{white}\bfseries Gain} & {\color{white}\bfseries 95\% CI}\\
\midrule
4B IT & Reference & +12.60 & [9.96,15.31]\\
4B IT & No reference & +12.67 & [10.13,15.34]\\
12B IT & Reference & +1.19 & [0.34,2.08]\\
12B IT & No reference & +2.13 & [1.29,3.13]\\
\bottomrule
\end{tabular}
\caption{Paired holistic-score differences, \pct{} minus baseline, on the matched 300-item subset.}
\label{tab:ci}\end{table}

Against 4B IT, \pct{} wins, ties, or loses on 214, 54, and 32 reference-assisted examples, respectively. Across rubric dimensions, its terminology mean increases from 75.60 to 90.42, and target-language adherence from 82.88 to 96.15. Terminology nevertheless remains its lowest reference-assisted dimension. These judgments suggest a useful focus for further bilingual error analysis; they do not establish the frequency of specific error categories. Complete rubric results by direction and target language are included in the appendix.

\section{Discussion}
The method combines three forms of supervision with different roles. Continued pretraining adapts language modeling to encyclopedic text. Multilingual answering requires selecting the information requested by a question and expressing it in a specified language; answer-text translation exposes the model to semantic counterparts in that same response space. Final bidirectional SFT then concentrates training on the translation relation evaluated at test time. Shared answer targets provide a concrete connection between the intermediate tasks, while the final stage uses a separate source-generation process.

The results show that this complete recipe produces a stronger translator on the recorded test set. They do not establish which component causes the gains. A controlled attribution study would keep the Wikipedia-adapted checkpoint, total training budget, final translation corpus, and inference prompts fixed while varying the intermediate answering/translation mixture. Evaluating the intermediate checkpoint would further distinguish improvements before and after specialization. These comparisons are not available here. The present contribution is a fully specified system and its translation evaluation, with corpus counts that make those future comparisons possible.

\section{Conclusion}
We presented \pct{}, a six-language translator adapted from Gemma 3 4B IT through Wikipedia continued pretraining, multilingual answering with answer-text translation, and bidirectional translation specialization. The central data design links answering and translation through multilingual answer targets, with 5.76 million cross-language QA records embedded in a 30.15-million-record intermediate mixture. On a 36,572-example synthetic test set, the final system improves all reported aggregate metrics over public Gemma 3 4B IT and 12B IT, including chrF++ gains in every direction. This study establishes a reproducible translation-focused adaptation recipe and a direction-level account of its strengths and remaining weaknesses.
\label{sec:mainend}
\clearpage
\section*{Limitations}
The test sources and references are synthetic, and the judge belongs to the same teacher-service family. Reference similarity and judge preferences may favor teacher-like outputs. The results therefore do not establish quality on human-authored benchmarks or out-of-domain text. The record-level split allows topic overlap; two source strings overlap after whitespace normalization, and the audit does not establish absence of semantic duplicates or pretraining exposure. Archived inference prompts differ across systems, so adaptation and prompting are not separated. Some baseline requests encountered context limits; their empty responses remain in the evaluation. COMET truncates long inputs, limiting the text considered by that metric. The judge subset contains only ten examples per direction. No human assessment, controlled mixture ablation, evaluation of intermediate checkpoints, or held-out QA evaluation is available. Training data have uneven language coverage, empty fields, and 1,961 conflicting answer-language directives. Internal validation sets are too small for reliable model selection. These limitations constrain both causal attribution and generalization of the reported gains.

\section*{Ethical Considerations}
Wikipedia-derived and synthetic data can reflect uneven representation across languages and topics. Generated translations may alter names, factual claims, or register; the model has not been validated for high-stakes translation. Data and model use must respect applicable Wikipedia attribution/share-alike obligations, upstream dataset and teacher-service terms, and Gemma terms. The archived exports do not recover complete article-level provenance, so this paper does not assert a blanket license over all text. We report existing training and generation artifacts without collecting new human annotations. Full directional reporting and retention of empty outputs help make performance differences across languages visible.

\bibliography{references}

\appendix
\section{Continued-pretraining corpus}
\label{app:ptdata}
The five distinct source files registered for phase 1 contain the raw text rows in Table~\ref{tab:ptrows}. Four Persian registrations refer to the same source file, so its rows are counted once in this table. The counts were measured from the available raw exports; they are neither unique-article counts nor the number of processed training sequences. The recorded training run reports 3,399,664 examples after preprocessing. No claim is made that the raw row sum alone reconstructs its tokenization, packing, or source repetition.
\begin{table}[t]\centering\small
\begin{tabular}{lr}\toprule
\rowcolor{TableHeaderDark}{\color{white}\bfseries Language} & {\color{white}\bfseries Raw text rows}\\\midrule
English & 6,406,711\\
Persian & 802,549\\
Arabic & 1,216,238\\
Chinese & 1,354,407\\
Egyptian Arabic & 1,615,156\\\midrule
Total across distinct source files & 11,395,061\\\bottomrule
\end{tabular}
\caption{Available phase-1 Wikipedia sources before training preprocessing. The registered mixture repeats the Persian source four times.}
\label{tab:ptrows}\end{table}

\section{Construction and inference details}
\label{app:construction}
\subsection{Multilingual answer inventory}
The phase-2 inventory begins with three question formats (general, standard, and title-derived QA). Records are pooled by format and task in the released dataset. Their combined initial row count is 2,025,730. Gemma 3 27B IT \citep{gemmateam2025gemma3} generates Persian-to-English and Persian-to-Arabic variants, followed by English-to-Chinese, English-to-Hebrew, and English-to-Spanish variants. Thus part of the inventory is English-pivoted. Equivalence of all generated answer variants has not been exhaustively checked by human annotators. Wikipedia context supports upstream example construction; the final training record supplies only an instruction, input text, and target response.

Question columns supply QA inputs and answer columns supply QA targets. For answer-text MT, both columns contain answer variants. Sampling fractions for QA depend on the source group, direction, and response variant. For answer-text MT, the fraction of available content retained for a direction is
\begin{equation}
 f(s,t)=\begin{cases}
 .70,&\{s,t\}=\{\mathrm{fa},\mathrm{en}\},\\
 .25,&\{s,t\}\cap\{\mathrm{fa},\mathrm{en}\}\ne\varnothing,\\
 .08,&\text{otherwise}.
 \end{cases}
\label{eq:legacy}
\end{equation}
Here $s,t$ are input and target languages and $s\ne t$. The first matching case applies. These fractions determine dataset construction; all retained answer-translation datasets are subsequently selected for phase 2. The construction records contain 10,695,112 MT rows; null filtering leaves 10,695,072. Across all tasks, summing retained construction rows reproduces 30,153,404 training records. The public release organizes these into 14 unified task configurations without dropping or resampling examples. Their internal split contains 30,153,102 training and 302 validation examples.

\subsection{System-instruction composition}
For cross-language complete-answer QA, the constructor samples a completeness directive and an answer-language directive, then combines them in randomized order with a sampled separator. Some same-language examples use only the completeness directive. Style-conditioned examples may add a formal or informal directive. The language in which a sentence is written is distinct from the answer language it requests. The construction uses random seed 42; this value controls reproducibility of sampling choices, not an additional modeling assumption. Prompt pools retain duplicate rows and their sampling mass.

\subsection{Synthetic translation generation}
Source generation uses a three-level topic taxonomy and attributes for genre, register, style, complexity, length, and optional linguistic/noise conditions. Stored metadata identifies DeepSeek-V3.1 \citep{deepseekai2025v31}; source generation uses temperature 0.5 and top-$p=0.9$. The translation service is addressed through \texttt{deepseek-chat}, without a recoverable immutable serving revision for every call. A different target language is selected for each source. Translation generation uses temperature zero and a 5,000-token completion budget. The teacher is asked to preserve content, names, numbers, formatting, and register; these instructions do not guarantee translation quality.

The aggregate contains 365,718 pairs. A shuffled 90/10 record split produces 329,146 raw train and 36,572 test pairs. No split seed is recorded. Language-label normalization, pair deduplication, and removal of empty/equal texts yield 329,137 forward and 329,137 reverse training records. The net reduction is nine underlying pairs; a per-filter attribution is unavailable. The final selected total is 658,274, with an internal train/validation split of 658,267/7. The stored split memberships are preserved throughout the retrospective evaluation.

\subsection{Archived inference}
\label{app:inference}
All three systems use temperature-zero decoding with a 5,000-token completion budget. \pct{} receives a short English direction instruction. The public Gemma models receive a longer English instruction specifying translation and formatting requirements, with quoted source text. The comparison therefore includes prompt differences. The full test contains 31 empty responses for Gemma 3 4B IT and none for the other two systems. Context-limit failures appear in the logs, so the empty responses cannot all be interpreted as linguistic failures. Empty strings and unsolicited wrappers remain in metric inputs.

\subsection{Judge rubric and paired intervals}
\label{app:uncertainty}
The reference-assisted judge sees source, candidate, and reference; the reference-free judge sees source and candidate. Both modes use a custom holistic 0--100 score and eight component scores, following the general model-based evaluation setting of \citet{kocmi-federmann-2023-large,kocmi-federmann-2023-gemba}. This is not a claim to reproduce GEMBA's exact protocol. Prompts request score caps of 10 for wrong-language output or untranslated copying, 40 for key mistranslations or hallucinations, and 60 for critical entity/number corruption. These are requested constraints, not verified error labels.

Let $J_{d,i}^{m,v}$ be the holistic judge score for system $m$, direction $d\in\Omega$, item $i\in\{1,\ldots,10\}$, and view $v\in\{\mathrm{ref},\mathrm{noref}\}$. For \pct{} (system $h$) and baseline $b$, the estimated score difference is
\begin{equation}
\widehat\Delta_v=\frac{1}{30}\sum_{d\in\Omega}\frac{1}{10}\sum_{i=1}^{10}
 (J_{d,i}^{h,v}-J_{d,i}^{b,v}).
\end{equation}
Each bootstrap replicate samples ten item indices with replacement within each direction, using the same sampled indices for both systems. We generate 10,000 replicates with seed 20260912 and use their 2.5th and 97.5th percentiles. These are paired, direction-stratified item-sampling intervals \citep{koehn-2004-statistical}.

\input{results/appendix_phase2.tex}
\begin{figure*}[t]\centering\includegraphics[width=\textwidth]{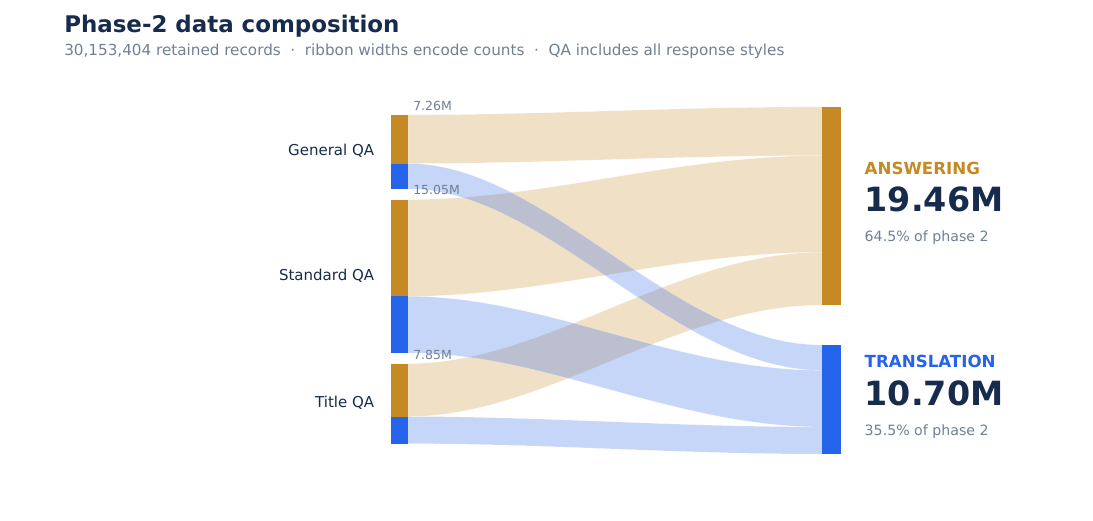}
\caption{Retained phase-2 records flowing from the three pooled question formats to answering and translation tasks. Ribbon widths encode record counts. Translation and several answering variants are jointly selected; source-row reuse across these tasks means the plotted counts are not independent documents.}\label{fig:composition}\end{figure*}
\begin{figure*}[t]\centering\includegraphics[width=.96\textwidth]{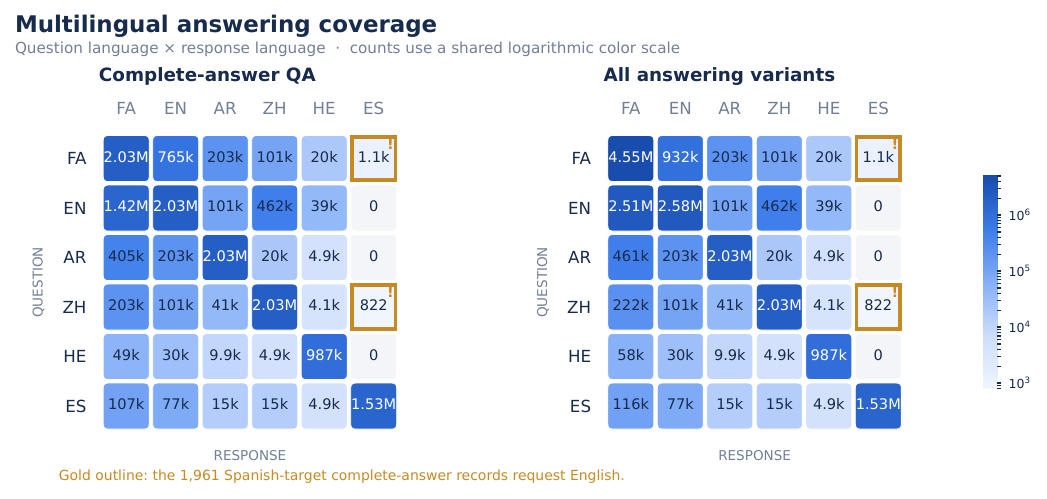}
\caption{Realized question/answer language coverage. Left: complete-answer QA. Right: all answering variants. Color uses a logarithmic count scale. Gold outlines mark the two Spanish-target cells whose complete-answer directives incorrectly request English (1,961 records). Zero cells indicate absent exported combinations.}\label{fig:qacounts}\end{figure*}

\section{Prompt sampling details}
\label{app:weights}
For instruction component $g$ (such as answer language or extent), let $\mathcal{P}_{g,p,t}$ be the list of sentences written in language $p$ for target language $t$. Let $\mathcal{A}_{g,t}$ contain the languages with available, nonempty pools. The builder normalizes configured nonnegative weights $w_{g,p}$ over $\mathcal{A}_{g,t}$, chooses $p$, and samples row index $j$ uniformly:
\begin{equation}
 P(p,j\mid g,t)=\frac{w_{g,p}}{\sum_{u\in\mathcal{A}_{g,t}}w_{g,u}}
 \frac{1}{|\mathcal{P}_{g,p,t}|}.
\label{eq:pool}
\end{equation}
Here $p\in\mathcal{A}_{g,t}$ and $1\leq j\leq|\mathcal{P}_{g,p,t}|$. The configured available weights have a positive sum. Repeated strings in a list therefore receive repeated sampling mass; uniformity is over rows, not distinct paraphrases. The prompt-generation scripts expand seed sentences with model-generated paraphrases, combine outputs, and substitute localized language names into target placeholders. Appendix~\ref{app:phase2} reports pool sizes and exact duplicate counts. Changing $p$ changes how a directive is expressed; changing $t$ changes the required response. Our core supervision concerns the latter.

\subsection{Final-phase translation instructions}
\label{sec:prompts}
For each direction, the constructor chooses a prompt-language pool, then uniformly samples a sentence from that pool. Its principal weights are
\begin{equation}
 \pi(p\mid s,t)=\begin{cases}
 .55,&p=t,\\ .30,&p=s,\\ w_p(s,t),&p\notin\{s,t\},
 \end{cases}
\end{equation}
The residual satisfies $\sum_{p\notin\{s,t\}}w_p=.15$. Remaining weights use a script-based heuristic: Persian/Arabic/Hebrew form an RTL group, English/Spanish a Latin group, and Chinese a CJK group. These are implementation groups, not genealogical language families. The allocation below specifies these weights. Empty pools are excluded and available weights renormalized by the sampler. Seed 42 controls the sampled construction choices; it does not retroactively seed source generation or the earlier raw split.

A final-phase translation record has the training schema \texttt{system}, \texttt{instruct}, \texttt{output}. Per-direction intermediate files retain explicit source/target labels, but the merger keeps the three training fields, drops missing values, and escapes media-like tags as text. Here $p$ denotes the language of the sampled translation instruction and supplies surface-form diversity. It does not add a loss that explicitly equates outputs across prompt languages, and our evaluation does not estimate such invariance.

Let $U=\Lset\setminus\{s,t\}$, $M\subseteq U$ contain languages sharing an implementation script group with $s$ or $t$, and $O=U\setminus M$. The constructor reserves 0.55 for $t$ and 0.30 for $s$. With residual $R=.15$, define
\begin{align}
 g&=\min(R/2,.01|O|),\\
 h&=\begin{cases}\min(.10,(R-g)/|M|),&|M|>0,\\0,&|M|=0,\end{cases}\\
 r_0&=R-g-|M|h.
\end{align}
For $p\in U$, the residual weight is
\begin{equation}
 w_p=\mathbf{1}[p\in O]\frac{g}{|O|}
 +\mathbf{1}[p\in M]h+\frac{r_0}{|U|},
\end{equation}
where a fraction on an empty group contributes zero. This matches the implementation's residual redistribution before rounding and final sampler normalization. For Persian-to-English, the unrounded weights in order (fa,en,es,he,zh,ar) are $(.30,.55,.046\overline6,.046\overline6,.01,.046\overline6)$. Here $\mathbf{1}[\cdot]$ is the indicator function. With $\mathcal{P}_{p,s,t}$ the nonempty prompt pool, the probability of selecting row $j$ from language pool $p$ is
\begin{equation}
 P(j,p\mid s,t)=\frac{\pi(p\mid s,t)}{|\mathcal{P}_{p,s,t}|},\quad 1\le j\le|\mathcal{P}_{p,s,t}|.
\end{equation}
If pools are unavailable, $\pi$ is renormalized over the available pools. The six-language assignment has full theoretical support, but observed coverage must be measured from instantiated prompts rather than inferred from the formula.

\subsection{Partition audit}
Streaming comparison of raw source strings finds zero exact train--test source overlaps. After collapsing whitespace, two test source texts match training sources, corresponding to 0.0055\% of the test partition; neither occurs in the 300-item judge subset. This is a narrow source-text audit: it does not prove absence of semantic duplicates, topic overlap, overlap with reverse inputs, or exposure during pretraining. The split is by record rather than by topic or Wikipedia article. We retain this distinction when interpreting generalization.

\section{Training and serialization}
\label{app:training}
\begin{figure*}[t]
\centering
\includegraphics[width=\textwidth]{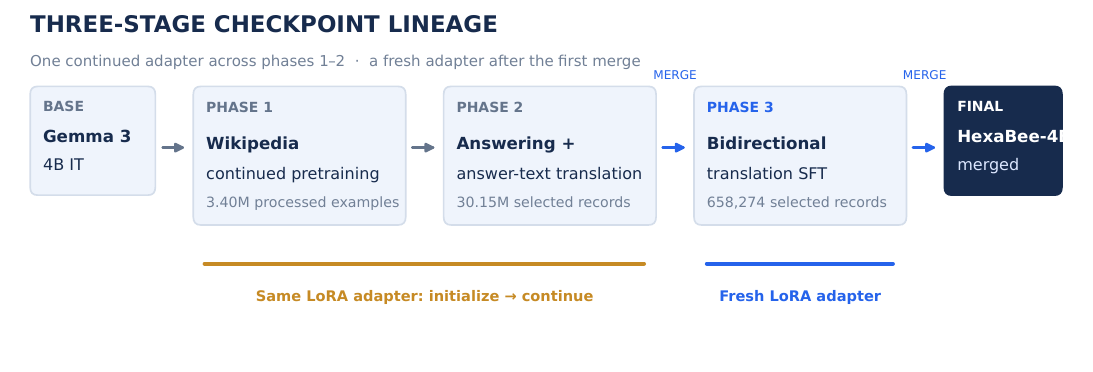}
\caption{Training lineage. The translation stage starts a new adapter on the merged cross-lingual model. Phase 2 combines system-directed answering with translations of multilingual answer texts; phase 3 specializes the merged model on independent synthetic translations.}
\label{fig:pipeline}
\end{figure*}
\begin{figure}[t]\centering\includegraphics[width=\linewidth]{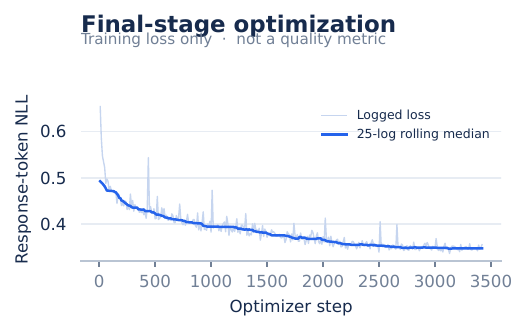}
\caption{Recorded response-token training loss during final translation adaptation, with a centered rolling median over up to 25 logged values for visual guidance. This is an optimization diagnostic, not a translation-quality or generalization metric.}\end{figure}

\begin{table*}[t]
\centering\small
\begin{tabular}{lrrr}
\toprule
\rowcolor{TableHeaderDark} {\color{white}\bfseries Setting} & {\color{white}\bfseries Wikipedia adaptation} & {\color{white}\bfseries Cross-lingual SFT} & {\color{white}\bfseries Translation SFT}\\
\midrule
Training examples & 3,399,664 & 30,153,102 & 658,267\\
Internal validation examples & Not established & 302 & 7\\
Epochs / optimizer steps & 1 / 13,280 & 1 / 6,544 & 1 / 3,429\\
CUDA processes & 4 & 6 & 2\\
Per-device batch / accumulation & 16 / 4 & 12 / 64 & 12 / 8\\
Effective batch & 256 & 4,608 & 192\\
Peak learning rate & $10^{-4}$ & $10^{-4}$ & $10^{-4}$\\
Schedule / warmup fraction & cosine / .05 & cosine / .05 & cosine / .05\\
Seed & 111 & 42 & 42\\
Runtime (hours) & 59.99 & 306.90 & 27.45\\
Adapter handling & new on public base & continue previous & new after merge\\
\bottomrule
\end{tabular}
\caption{Executed settings from training logs and generated model cards. Counts have stage-specific semantics and should not be added as unique documents. Current nearby YAMLs may describe other runs; logged values take precedence.}
\label{tab:stages}
\end{table*}
The final stage uses PEFT 0.17.1 \citep{huggingface2025peft}, Transformers 4.55.4 \citep{wolf-etal-2020-transformers}, PyTorch 2.7.1+cu126, Datasets 3.6.0, Tokenizers 0.21.1, and DeepSpeed 0.16.9 with ZeRO stage 0. AdamW \citep{loshchilov2019decoupled} parameters are $\beta_1=.9$, $\beta_2=.999$, and $\epsilon=10^{-8}$. Training uses 4,096-token sequences, response-only supervision, checkpoints every 600 steps, and internal evaluation every 200 steps. The final adapter has no bias adaptation, DoRA, or rank-stabilized LoRA. The saved configuration of the continued cross-lingual adapter also uses dropout .005 despite a different value in the nearby YAML.

For a text-only example, the serialized prefix is shown schematically below; literal line breaks separate fields:
{\small
\begin{verbatim}
<bos><start_of_turn>user
[system instruction]

[source text]<end_of_turn>
<start_of_turn>model
[target text]<end_of_turn>
\end{verbatim}
}
\begin{equation}
 W'_j=W_j+\frac{\alpha}{r}B_jA_j,
\end{equation}
where $W_j\in\mathbb{R}^{d_{o,j}\times d_{i,j}}$ is projection $j$ with input dimension $d_{i,j}$ and output dimension $d_{o,j}$, $A_j\in\mathbb{R}^{r\times d_{i,j}}$ and $B_j\in\mathbb{R}^{d_{o,j}\times r}$ are trainable factors, $r$ is adapter rank, and $\alpha$ is its scale \citep{hu2022lora}. The saved adapters use $r=256$, $\alpha=512$, and dropout 0.005. These standard LoRA settings are implementation choices; the method's supervision relations are given in the main text.

The model's 34 text layers have hidden size 2,560, eight query heads, four key/value heads, and MLP intermediate size 10,240. The multimodal checkpoint retains its vision components, although final-stage adaptation affects only text projections. The nominal 4B size identifies the Gemma model family. For the adapted projections, the parameter count is
\begin{equation}
 N_{\mathrm{LoRA}}=r\sum_j(d_{i,j}+d_{o,j})=476{,}839{,}936.
\end{equation}
Each layer contributes two $2{,}560\times2{,}048$ attention projections, two $2{,}560\times1{,}024$ projections, and three $2{,}560\times10{,}240$ MLP projections, up to transposition. Summing their input/output dimensions over 34 layers verifies the adapter count.

\section{Metrics and results by language}
\label{app:metrics}
Language codes are ar (Arabic), zh (Chinese), en (English), he (Hebrew), fa (Persian), and es (Spanish). Direction tables keep source and target distinct. Target summaries give each of the five incoming directions equal weight; they are direction-macro summaries, not recomputed pooled corpora. Judge dimension scores remain on a 0--100 scale and are not calibrated accuracy percentages. All raw model responses are retained for scoring, including quotation wrappers. The machine-readable artifacts include metric signatures and sample counts.
Writing $M_{s,t}$ for the score of direction $(s,t)$, the target-language summary is
\begin{equation}
 M_{\to t}=\frac{1}{5}\sum_{s\in\Lset\setminus\{t\}} M_{s,t}.
\end{equation}
Pooled corpus chrF++ is 60.77 for Gemma 3 4B IT, 71.13 for Gemma 3 12B IT, and 77.30 for \pct{}. These differ from the direction-macro values in Table~\ref{tab:main}.
Let $\mathcal{G}$ index the configured n-gram types, each defined by a unit (character or word) and an order. For $g\in\mathcal{G}$, let $H_g$, $R_g$, and $C_g$ denote corpus hypothesis, reference, and matched counts. With $E=\{g\in\mathcal{G}:H_g>0, R_g>0\}$, the effective-order averages are
\begin{equation}
 P=\frac{1}{|E|}\sum_{g\in E}\frac{C_g}{H_g},\qquad
 R=\frac{1}{|E|}\sum_{g\in E}\frac{C_g}{R_g}.
\end{equation}
For chrF++, orders include character 1--6 and word 1--2; chrF uses only the character orders. SacreBLEU then computes
\begin{equation}
 \mathrm{chrF}_{\beta}=100(1+\beta^2)\frac{PR}{\beta^2P+R},\quad\beta=2,
\end{equation}
with a zero score for an empty effective set or zero denominator. It aggregates corpus counts before the harmonic combination, rather than averaging sentence scores.
BLEU is computed from clipped corpus precisions $p_n$ and total hypothesis/reference lengths $c,r$:
\begin{equation}
 \mathrm{BLEU}=100\,\mathrm{BP}\exp\!\left(\frac14\sum_{n=1}^{4}\log p_n\right),
\end{equation}
where $\mathrm{BP}=\min(1,\exp(1-r/c))$ for $c>0$, with SacreBLEU's exponential smoothing for zero matches. TER uses the Tercom-style edit search:
\begin{equation}
 \mathrm{TER}=100\frac{N_{\mathrm{sub}}+N_{\mathrm{del}}+N_{\mathrm{ins}}+N_{\mathrm{shift}}}{N_{\mathrm{ref}}}.
\end{equation}
Here $N_{\mathrm{sub}}$, $N_{\mathrm{del}}$, $N_{\mathrm{ins}}$, and $N_{\mathrm{shift}}$ count substitutions, deletions, insertions, and shifts; $N_{\mathrm{ref}}$ is the total reference-token count. This search approximates the edit sequence with shifts; the score should not be described as a certified global minimum of all possible edit programs. Corpus and macro aggregation are kept distinct for both metrics.

Let $f_\omega$ be the COMET regression model with fixed checkpoint parameters $\omega$ and $\tau_{512}$ its tokenizer/encoder truncation operation. For $N$ examples, let $x_i$, $\hat y_i$, and $y_i$ denote source, candidate, and reference text, respectively. The scaled corpus mean is
\begin{equation}
 \frac{100}{N}\sum_{i=1}^{N}f_\omega\!\left(\tau_{512}(x_i),\tau_{512}(\hat y_i),\tau_{512}(y_i)\right).
\end{equation}
Direction-macro COMET instead averages the 30 direction means. Scaling by 100 does not turn the regression output into an accuracy percentage.

\begin{figure*}[t]\centering\includegraphics[width=.96\textwidth]{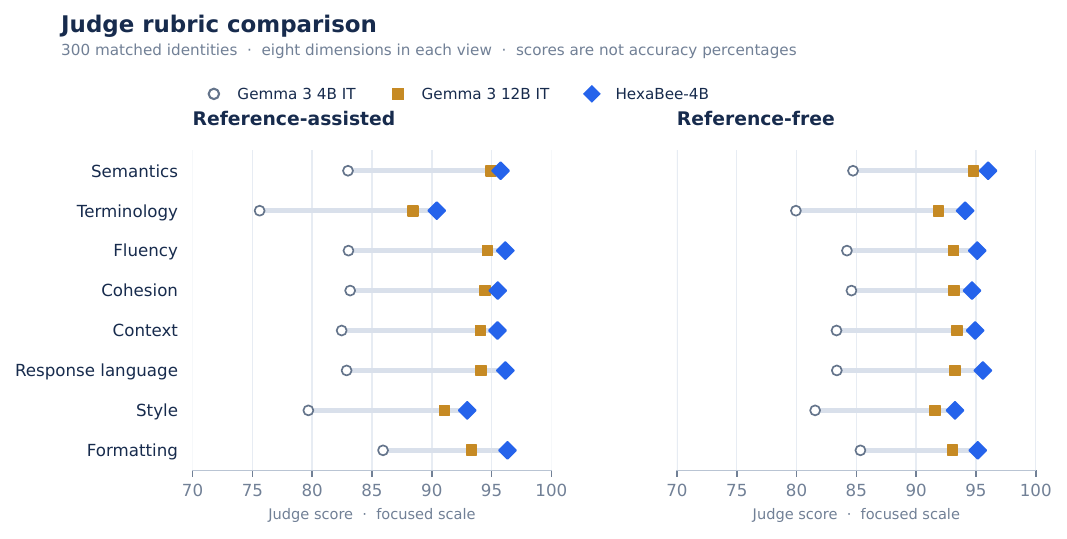}
\caption{All eight judge dimensions, averaged over the same 300 identities. Both views compare the same three systems; they remain judgments from one evaluator and are not independent human assessments.}\label{fig:rubric}\end{figure*}
\begin{figure*}[t]\centering\includegraphics[width=.95\textwidth]{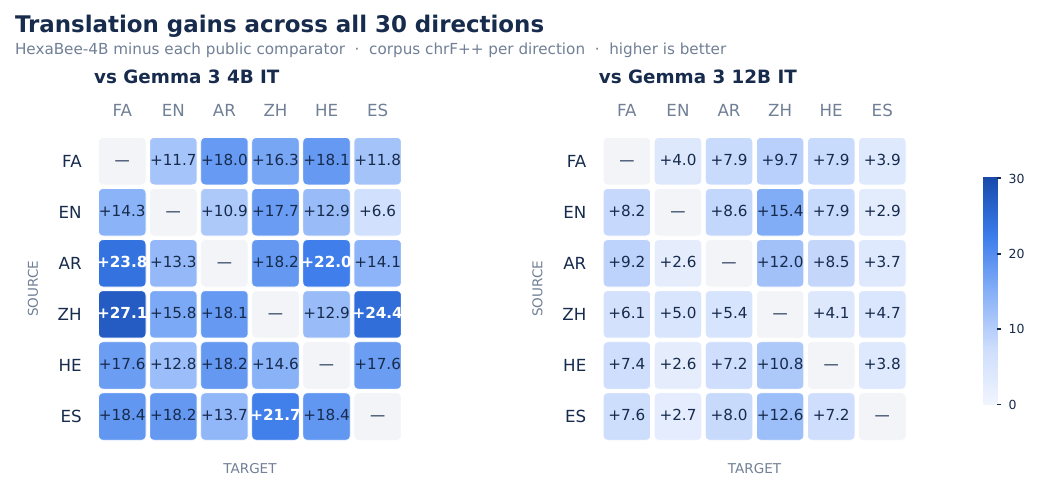}
\caption{Full-test chrF++ gains of \pct{} over both public comparison systems across all 30 source--target directions.}\label{fig:gains}\end{figure*}
\input{results/appendix_metrics.tex}

\section{Evaluation integrity and reproducibility}
The parser used truth-value fallback for the score field, which turned some explicit zero scores into nulls in intermediate records. We reparse the original judge JSON and recover 11 reference-view and 13 reference-free zeros for the 4B baseline, plus one reference-free zero for the 12B baseline. All 300 items then have valid scores in both views. Dropping these entries would inflate baseline means. The reparsed means agree with the final flattened score files.

Content identity is the SHA-256 hash of a canonical JSON list containing source language, target language, source text, and reference. We require equality of identity sets before comparing full-test files and before pairing judge scores. File row indices are not stable identities: asynchronous processing and sampling changed order across models. For judge recovery, an explicit JSON field \texttt{score: 0} remains zero rather than being treated as a missing value.

Our audit scripts are read-only with respect to the original training and evaluation artifacts. They preserve null/empty outputs, pin metric settings, save directional sufficient statistics where applicable, and produce content-hash IDs without copying raw evaluation text into the paper package. The local evidence directory records the actual training/merge paths and sanitized snapshots for author verification; those private paths are omitted from this release. The paper reports a retrospective analysis of existing translation outputs and newly computed metrics, not new training runs or newly collected human labels.
\end{document}

%% file: results/phase2_main_table.tex
\begin{table}[t]\centering\small
\begin{tabular}{lr}\toprule
\rowcolor{TableHeaderDark} {\color{white}\bfseries Phase-2 supervision} & {\color{white}\bfseries Records}\\\midrule
Same-language answering & 13,698,526\\
Cross-language answering & 5,759,806\\
Answer-text translation & 10,695,072\\
\midrule Total & 30,153,404\\\bottomrule
\end{tabular}
\caption{Retained phase-2 supervision. Same/cross refers to question and answer language. Response-style and direction breakdowns are given in the appendix.}
\label{tab:phase2}\end{table}

%% file: results/main_comparison.tex
\begin{table*}[t]\centering\small
\setlength{\tabcolsep}{2.8pt}
\renewcommand{\arraystretch}{1.08}
\begin{tabular}{lrrrrrrr}
\toprule
\rowcolor{TableHeaderDark} {\color{white}\bfseries System} & {\color{white}\bfseries BLEU} & {\color{white}\bfseries chrF} & {\color{white}\bfseries chrF++} & {\color{white}\bfseries TER$\downarrow$} & {\color{white}\bfseries COMET} & {\color{white}\bfseries Ref. J.} & {\color{white}\bfseries No-ref. J.}\\
\midrule
4B IT & 40.70 & 58.91 & 57.11 & 51.40 & 85.88 & 80.32 & 81.43\\
12B IT & \underline{50.54} & \underline{68.70} & \underline{66.83} & \underline{38.89} & \underline{88.20} & \underline{91.73} & \underline{91.97}\\
\rowcolor{PCTTint} \pct{} & \textbf{62.03} & \textbf{75.08} & \textbf{73.75} & \textbf{30.05} & \textbf{89.86} & \textbf{92.92} & \textbf{94.10}\\
\bottomrule\end{tabular}
\caption{Full-test direction-macro automatic metrics ($N=36{,}572$) and holistic judge means on 300 matched items. COMET is scaled by 100; TER is lower-is-better. Bold marks the best displayed score and underlining the second distinct score. Each system uses its recorded inference prompt.}
\label{tab:main}
\end{table*}

%% file: results/appendix_phase2.tex
\section{Phase-2 corpus composition}\label{app:phase2}
The following counts pool all retained records by question format and task. Summing construction records with non-null training fields exactly reproduces the training mixture. The released data have 14 unified task configurations; merging source partitions changes only packaging, not the examples. Source/target labels for answering tasks are independently checked against the notebook column policies. Counts describe records, not unique articles or independent question concepts.
\begin{table*}[t]\centering\small
\begin{tabular}{lrrrrrr}\toprule
\rowcolor{TableHeader} Subcorpus & Complete QA & Moderate QA & Formal QA & Informal QA & Translation & Total\\\midrule
General QA & 3,952,557 & 208,364 & 330,194 & 284,650 & 2,481,368 & 7,257,133\\
Standard QA & 7,240,185 & 300,875 & 1,540,222 & 411,033 & 5,554,592 & 15,046,907\\
Title QA & 3,835,308 & 0 & 646,570 & 708,374 & 2,659,112 & 7,849,364\\
\bottomrule\end{tabular}
\caption{All selected phase-2 data by question format and task, with source partitions pooled. A zero denotes an absent task export.}\end{table*}
\begin{table}[t]\centering\scriptsize
\setlength{\tabcolsep}{3pt}
\begin{tabular}{lrrrrrr}\toprule
\rowcolor{TableHeader} $s\backslash t$ & fa & en & ar & zh & he & es\\\midrule
fa & 2,025,720 & 764,747 & 202,573 & 101,287 & 19,744 & 1,139\\
en & 1,418,003 & 2,025,730 & 101,287 & 462,076 & 39,489 & 0\\
ar & 405,143 & 202,573 & 2,025,730 & 20,258 & 4,936 & 0\\
zh & 202,572 & 101,287 & 40,514 & 2,025,730 & 4,114 & 822\\
he & 49,362 & 29,616 & 9,873 & 4,936 & 987,223 & 0\\
es & 107,249 & 76,607 & 15,322 & 15,322 & 4,936 & 1,532,130\\
\bottomrule\end{tabular}
\caption{Complete QA records by input language (rows) and response language (columns). Diagonal cells are same-language answering; translation excludes identity directions.}\end{table}
\begin{table}[t]\centering\scriptsize
\setlength{\tabcolsep}{3pt}
\begin{tabular}{lrrrrrr}\toprule
\rowcolor{TableHeader} $s\backslash t$ & fa & en & ar & zh & he & es\\\midrule
fa & 4,545,437 & 931,711 & 202,573 & 101,287 & 19,744 & 1,139\\
en & 2,513,228 & 2,582,276 & 101,287 & 462,076 & 39,489 & 0\\
ar & 460,798 & 202,573 & 2,025,730 & 20,258 & 4,936 & 0\\
zh & 222,051 & 101,287 & 40,514 & 2,025,730 & 4,114 & 822\\
he & 57,710 & 29,616 & 9,873 & 4,936 & 987,223 & 0\\
es & 115,597 & 76,607 & 15,322 & 15,322 & 4,936 & 1,532,130\\
\bottomrule\end{tabular}
\caption{All QA response variants (complete, moderate, formal, informal) records by input language (rows) and response language (columns). Diagonal cells are same-language answering; translation excludes identity directions.}\end{table}
\begin{table}[t]\centering\scriptsize
\setlength{\tabcolsep}{3pt}
\begin{tabular}{lrrrrrr}\toprule
\rowcolor{TableHeader} $s\backslash t$ & fa & en & ar & zh & he & es\\\midrule
fa & 0 & 1,418,003 & 506,429 & 506,429 & 246,806 & 383,029\\
en & 1,418,003 & 0 & 506,433 & 506,433 & 246,806 & 383,033\\
ar & 506,429 & 506,433 & 0 & 162,059 & 78,978 & 122,571\\
zh & 506,429 & 506,433 & 162,059 & 0 & 78,978 & 122,571\\
he & 246,806 & 246,806 & 78,978 & 78,978 & 0 & 78,978\\
es & 383,029 & 383,033 & 122,571 & 122,571 & 78,978 & 0\\
\bottomrule\end{tabular}
\caption{Answer-text MT records by input language (rows) and response language (columns). Diagonal cells are same-language answering; translation excludes identity directions.}\end{table}
\paragraph{Construction quality audit.} The merger removes null fields but preserves empty strings. The final exports contain 6,742 empty input fields and 6,043 empty response fields (field counts can overlap within a record); no system field is empty. These entries are included in the recorded export counts. The raw construction-to-export reduction is 69 records, including 40 translation records. Two complete-answer exports contain 822 Chinese-to-Spanish and 1,139 Persian-to-Spanish records whose language-directive pools request English. Both conflicts are verified in stored examples. They account for all 1,961 cross-language Spanish-target complete-answer records. Consequently Spanish cross-language answering with a correctly specified target is not established by this subset, although Spanish is present in same-language QA and translation supervision. The historical model is evaluated as trained; the audit does not retroactively repair its supervision.
\begin{table}[t]\centering\small
\begin{tabular}{lrrrr}\toprule
 & \multicolumn{2}{c}{Answer-language} & \multicolumn{2}{c}{Completeness}\\
Prompt language & Rows & Unique & Rows & Unique\\\midrule
fa & 1302 & 1110 & 1076 & 836\\
en & 1302 & 944 & 1000 & 759\\
ar & 1060 & 817 & 1032 & 792\\
zh & 976 & 755 & 1024 & 784\\
he & 914 & 691 & 1080 & 840\\
es & -- & -- & 1009 & 728\\
\bottomrule\end{tabular}
\caption{Representative prompt pools before record sampling. Counts use the pool whose target-language label matches its prompt language; completeness text does not itself specify the answer language. Unique counts deduplicate exact sentence strings. Repeated rows remain in the sampler. A dash indicates no such pool.}\label{tab:pools}\end{table}

%% file: results/appendix_metrics.tex
\begin{table*}[t]\centering\small
\setlength{\tabcolsep}{2.8pt}
\renewcommand{\arraystretch}{1.08}
\begin{tabular}{llrrrrrrrr}
\toprule
\rowcolor{TableHeaderDark} {\color{white}\bfseries Target} & {\color{white}\bfseries System} & {\color{white}\bfseries $N$} & {\color{white}\bfseries BLEU} & {\color{white}\bfseries chrF} & {\color{white}\bfseries chrF++} & {\color{white}\bfseries TER$\downarrow$} & {\color{white}\bfseries COMET} & {\color{white}\bfseries Ref. J.} & {\color{white}\bfseries No-ref. J.}\\
\midrule
All & 4B IT & 36,572 & 40.70 & 58.91 & 57.11 & 51.40 & 85.88 & 80.32 & 81.43\\
 & 12B IT & 36,572 & \underline{50.54} & \underline{68.70} & \underline{66.83} & \underline{38.89} & \underline{88.20} & \underline{91.73} & \underline{91.97}\\
\rowcolor{PCTTint}  & \pct{} & 36,572 & \textbf{62.03} & \textbf{75.08} & \textbf{73.75} & \textbf{30.05} & \textbf{89.86} & \textbf{92.92} & \textbf{94.10}\\
\addlinespace[2pt]
ar & 4B IT & 6,120 & 34.09 & 62.25 & 59.33 & 56.99 & 86.64 & 82.56 & 81.60\\
 & 12B IT & 6,120 & \underline{43.14} & \underline{70.43} & \underline{67.69} & \underline{44.60} & \underline{88.52} & \underline{89.12} & \underline{89.78}\\
\rowcolor{PCTTint}  & \pct{} & 6,120 & \textbf{56.17} & \textbf{77.09} & \textbf{75.10} & \textbf{33.62} & \textbf{90.17} & \textbf{92.72} & \textbf{93.60}\\
\addlinespace[2pt]
zh & 4B IT & 6,135 & 35.22 & 31.87 & 30.97 & 60.19 & 84.52 & 75.22 & 76.00\\
 & 12B IT & 6,135 & \underline{40.15} & \underline{38.40} & \underline{36.57} & \underline{54.28} & \underline{87.01} & \underline{89.92} & \underline{90.82}\\
\rowcolor{PCTTint}  & \pct{} & 6,135 & \textbf{54.78} & \textbf{50.31} & \textbf{48.67} & \textbf{38.68} & \textbf{90.00} & \textbf{91.82} & \textbf{92.60}\\
\addlinespace[2pt]
en & 4B IT & 6,010 & 55.28 & 74.13 & 72.40 & 36.35 & 87.40 & 86.26 & 86.84\\
 & 12B IT & 6,010 & \underline{67.16} & \underline{84.77} & \underline{83.36} & \underline{22.59} & \underline{88.90} & \textbf{94.78} & \underline{94.42}\\
\rowcolor{PCTTint}  & \pct{} & 6,010 & \textbf{75.06} & \textbf{87.68} & \textbf{86.76} & \textbf{17.00} & \textbf{89.83} & \underline{93.52} & \textbf{94.66}\\
\addlinespace[2pt]
he & 4B IT & 6,196 & 31.94 & 54.98 & 53.02 & 59.27 & 85.87 & 76.00 & 78.44\\
 & 12B IT & 6,196 & \underline{43.32} & \underline{64.53} & \underline{62.78} & \underline{46.14} & \underline{88.17} & \underline{91.08} & \underline{91.40}\\
\rowcolor{PCTTint}  & \pct{} & 6,196 & \textbf{54.71} & \textbf{70.90} & \textbf{69.90} & \textbf{42.17} & \textbf{89.68} & \textbf{93.16} & \textbf{94.30}\\
\addlinespace[2pt]
fa & 4B IT & 6,069 & 32.44 & 57.57 & 55.67 & 59.16 & 84.81 & 76.86 & 77.86\\
 & 12B IT & 6,069 & \underline{43.61} & \underline{70.24} & \underline{68.22} & \underline{41.59} & \underline{88.32} & \underline{91.04} & \underline{92.60}\\
\rowcolor{PCTTint}  & \pct{} & 6,069 & \textbf{57.43} & \textbf{77.34} & \textbf{75.90} & \textbf{31.68} & \textbf{89.91} & \textbf{92.06} & \textbf{94.34}\\
\addlinespace[2pt]
es & 4B IT & 6,042 & 55.21 & 72.69 & 71.24 & 36.44 & 86.02 & 85.04 & 87.86\\
 & 12B IT & 6,042 & \underline{65.85} & \underline{83.83} & \underline{82.35} & \underline{24.13} & \underline{88.29} & \textbf{94.44} & \underline{92.82}\\
\rowcolor{PCTTint}  & \pct{} & 6,042 & \textbf{74.04} & \textbf{87.18} & \textbf{86.16} & \textbf{17.13} & \textbf{89.57} & \underline{94.24} & \textbf{95.12}\\
\bottomrule\end{tabular}
\caption{Overall and target-language comparisons. Automatic metrics are direction-macro averages on the full test set; holistic judge means use 300 matched items overall or 50 per target. COMET is scaled by 100; TER is lower-is-better. Bold marks the best displayed score and underlining the second distinct displayed score within each group; ties share rank. The adapted model is tinted for easy comparison. These recorded systems use different inference prompts.}
\label{tab:auto_all}
\end{table*}
\begin{table*}[t]\centering\scriptsize
\setlength{\tabcolsep}{2.8pt}
\renewcommand{\arraystretch}{1.08}
\begin{tabular}{llrrrrrrrrrrr}
\toprule
\rowcolor{TableHeaderDark} {\color{white}\bfseries Direction} & {\color{white}\bfseries System} & {\color{white}\bfseries $N$} & {\color{white}\bfseries BLEU} & {\color{white}\bfseries chrF} & {\color{white}\bfseries chrF++} & {\color{white}\bfseries TER$\downarrow$} & {\color{white}\bfseries COMET} & {\color{white}\bfseries Ref. J.} & {\color{white}\bfseries No-ref. J.} & {\color{white}\bfseries chrF++$_{10}$} & {\color{white}\bfseries TER$_{10}\downarrow$} & {\color{white}\bfseries COMET$_{10}$}\\
\midrule
$zh\!\to\!ar$ & 4B IT & 1229 & 23.68 & 51.49 & 48.28 & 80.63 & 84.07 & 72.60 & 71.60 & 37.49 & 100.59 & 80.35\\
 & 12B IT & 1229 & \underline{32.53} & \underline{64.44} & \underline{60.91} & \underline{61.20} & \underline{87.91} & \textbf{91.60} & \underline{92.50} & \underline{62.38} & \underline{53.76} & \underline{87.21}\\
\rowcolor{PCTTint}  & \pct{} & 1229 & \textbf{42.46} & \textbf{69.06} & \textbf{66.34} & \textbf{57.30} & \textbf{89.15} & \underline{90.40} & \textbf{93.00} & \textbf{71.36} & \textbf{38.78} & \textbf{88.68}\\
\addlinespace[2pt]
$en\!\to\!ar$ & 4B IT & 1213 & 37.50 & 66.77 & 63.77 & 51.70 & \underline{87.99} & \underline{86.50} & \underline{85.20} & \underline{65.44} & \underline{44.06} & \underline{88.04}\\
 & 12B IT & 1213 & \underline{42.30} & \underline{68.56} & \underline{66.09} & \underline{47.70} & 87.47 & 81.10 & 82.20 & 57.25 & 48.09 & 85.10\\
\rowcolor{PCTTint}  & \pct{} & 1213 & \textbf{55.60} & \textbf{76.61} & \textbf{74.66} & \textbf{35.72} & \textbf{90.06} & \textbf{93.40} & \textbf{94.10} & \textbf{76.86} & \textbf{27.55} & \textbf{90.48}\\
\addlinespace[2pt]
$he\!\to\!ar$ & 4B IT & 1230 & 39.08 & 63.41 & 61.23 & 48.02 & 86.42 & 78.30 & 74.70 & 50.31 & 52.36 & 80.29\\
 & 12B IT & 1230 & \underline{51.01} & \underline{74.60} & \underline{72.25} & \underline{32.19} & \underline{89.30} & \underline{88.40} & \underline{87.90} & \underline{64.18} & \underline{38.88} & \underline{84.90}\\
\rowcolor{PCTTint}  & \pct{} & 1230 & \textbf{63.65} & \textbf{80.99} & \textbf{79.44} & \textbf{23.26} & \textbf{90.59} & \textbf{92.60} & \textbf{93.60} & \textbf{69.00} & \textbf{31.11} & \textbf{86.10}\\
\addlinespace[2pt]
$fa\!\to\!ar$ & 4B IT & 1228 & 32.88 & 63.15 & 60.05 & 56.62 & 87.51 & 88.70 & 89.80 & 68.97 & 38.75 & 89.71\\
 & 12B IT & 1228 & \underline{46.62} & \underline{72.69} & \underline{70.14} & \underline{38.53} & \underline{89.75} & \underline{92.60} & \underline{93.40} & \underline{73.76} & \underline{31.76} & \underline{90.41}\\
\rowcolor{PCTTint}  & \pct{} & 1228 & \textbf{61.00} & \textbf{79.59} & \textbf{78.00} & \textbf{24.43} & \textbf{91.27} & \textbf{94.40} & \textbf{95.10} & \textbf{81.87} & \textbf{20.81} & \textbf{91.55}\\
\addlinespace[2pt]
$es\!\to\!ar$ & 4B IT & 1220 & 37.28 & 66.41 & 63.33 & 47.96 & 87.21 & 86.70 & 86.70 & 64.12 & 44.99 & 88.36\\
 & 12B IT & 1220 & \underline{43.22} & \underline{71.87} & \underline{69.06} & \underline{43.37} & \underline{88.15} & \underline{91.90} & \textbf{92.90} & \underline{69.86} & \underline{37.83} & \underline{88.93}\\
\rowcolor{PCTTint}  & \pct{} & 1220 & \textbf{58.14} & \textbf{79.19} & \textbf{77.07} & \textbf{27.40} & \textbf{89.78} & \textbf{92.80} & \underline{92.20} & \textbf{76.58} & \textbf{27.15} & \textbf{89.85}\\
\bottomrule\end{tabular}
\caption{All systems translating into Arabic. Columns through COMET use the full direction-specific test set ($N$). The last 5 columns use the ten shared judge-subset items per direction. Bold/underline mark first/second within the same direction and metric; TER is lower-is-better.}
\end{table*}
\begin{table*}[t]\centering\scriptsize
\setlength{\tabcolsep}{2.8pt}
\renewcommand{\arraystretch}{1.08}
\begin{tabular}{llrrrrrrrrrrr}
\toprule
\rowcolor{TableHeaderDark} {\color{white}\bfseries Direction} & {\color{white}\bfseries System} & {\color{white}\bfseries $N$} & {\color{white}\bfseries BLEU} & {\color{white}\bfseries chrF} & {\color{white}\bfseries chrF++} & {\color{white}\bfseries TER$\downarrow$} & {\color{white}\bfseries COMET} & {\color{white}\bfseries Ref. J.} & {\color{white}\bfseries No-ref. J.} & {\color{white}\bfseries chrF++$_{10}$} & {\color{white}\bfseries TER$_{10}\downarrow$} & {\color{white}\bfseries COMET$_{10}$}\\
\midrule
$ar\!\to\!zh$ & 4B IT & 1201 & 36.56 & 32.17 & 30.93 & 58.04 & 83.81 & 73.60 & 73.90 & 22.92 & 62.38 & 81.89\\
 & 12B IT & 1201 & \underline{42.36} & \underline{39.84} & \underline{37.13} & \underline{50.06} & \underline{86.94} & \underline{91.80} & \underline{91.90} & \underline{30.13} & \underline{51.75} & \underline{87.06}\\
\rowcolor{PCTTint}  & \pct{} & 1201 & \textbf{57.09} & \textbf{51.82} & \textbf{49.12} & \textbf{35.01} & \textbf{89.69} & \textbf{93.70} & \textbf{94.20} & \textbf{40.72} & \textbf{35.73} & \textbf{89.56}\\
\addlinespace[2pt]
$en\!\to\!zh$ & 4B IT & 1276 & 30.42 & 27.90 & 27.56 & 69.42 & 83.44 & 82.30 & 82.80 & 29.76 & 49.49 & 83.68\\
 & 12B IT & 1276 & \underline{32.77} & \underline{31.48} & \underline{29.94} & \underline{67.91} & \underline{85.07} & \underline{91.80} & \underline{93.50} & \underline{30.05} & \underline{44.73} & \underline{87.02}\\
\rowcolor{PCTTint}  & \pct{} & 1276 & \textbf{50.33} & \textbf{45.82} & \textbf{45.30} & \textbf{45.39} & \textbf{90.02} & \textbf{93.00} & \textbf{93.60} & \textbf{44.30} & \textbf{31.80} & \textbf{89.95}\\
\addlinespace[2pt]
$he\!\to\!zh$ & 4B IT & 1202 & 41.02 & 38.25 & 37.22 & 51.04 & 86.17 & 76.10 & 77.20 & 29.02 & 66.88 & 84.29\\
 & 12B IT & 1202 & \underline{44.57} & \underline{42.44} & \underline{40.97} & \underline{45.95} & \underline{87.67} & \underline{93.00} & \underline{93.50} & \underline{38.94} & \underline{52.52} & \underline{87.92}\\
\rowcolor{PCTTint}  & \pct{} & 1202 & \textbf{58.14} & \textbf{53.41} & \textbf{51.80} & \textbf{33.49} & \textbf{89.75} & \textbf{94.20} & \textbf{94.30} & \textbf{50.28} & \textbf{34.08} & \textbf{90.12}\\
\addlinespace[2pt]
$fa\!\to\!zh$ & 4B IT & 1251 & 38.95 & 35.74 & 34.01 & 53.12 & 86.19 & 74.70 & 75.00 & 37.67 & 48.16 & 83.33\\
 & 12B IT & 1251 & \underline{45.23} & \underline{42.86} & \underline{40.58} & \underline{44.42} & \underline{88.67} & \underline{92.10} & \underline{90.50} & \underline{46.31} & \underline{34.15} & \underline{89.23}\\
\rowcolor{PCTTint}  & \pct{} & 1251 & \textbf{57.23} & \textbf{52.86} & \textbf{50.30} & \textbf{34.91} & \textbf{90.41} & \textbf{94.40} & \textbf{94.60} & \textbf{58.36} & \textbf{23.84} & \textbf{91.03}\\
\addlinespace[2pt]
$es\!\to\!zh$ & 4B IT & 1205 & 29.14 & 25.27 & 25.15 & 69.33 & 83.01 & 69.40 & 71.10 & 13.03 & 71.67 & 79.39\\
 & 12B IT & 1205 & \underline{35.84} & \underline{35.37} & \underline{34.24} & \underline{63.07} & \underline{86.73} & \underline{80.90} & \underline{84.70} & \underline{17.67} & \underline{62.46} & \underline{82.49}\\
\rowcolor{PCTTint}  & \pct{} & 1205 & \textbf{51.11} & \textbf{47.63} & \textbf{46.84} & \textbf{44.61} & \textbf{90.14} & \textbf{83.80} & \textbf{86.30} & \textbf{24.63} & \textbf{44.99} & \textbf{85.80}\\
\bottomrule\end{tabular}
\caption{All systems translating into Chinese. Columns through COMET use the full direction-specific test set ($N$). The last 5 columns use the ten shared judge-subset items per direction. Bold/underline mark first/second within the same direction and metric; TER is lower-is-better.}
\end{table*}
\begin{table*}[t]\centering\scriptsize
\setlength{\tabcolsep}{2.8pt}
\renewcommand{\arraystretch}{1.08}
\begin{tabular}{llrrrrrrrrrrr}
\toprule
\rowcolor{TableHeaderDark} {\color{white}\bfseries Direction} & {\color{white}\bfseries System} & {\color{white}\bfseries $N$} & {\color{white}\bfseries BLEU} & {\color{white}\bfseries chrF} & {\color{white}\bfseries chrF++} & {\color{white}\bfseries TER$\downarrow$} & {\color{white}\bfseries COMET} & {\color{white}\bfseries Ref. J.} & {\color{white}\bfseries No-ref. J.} & {\color{white}\bfseries chrF++$_{10}$} & {\color{white}\bfseries TER$_{10}\downarrow$} & {\color{white}\bfseries COMET$_{10}$}\\
\midrule
$ar\!\to\!en$ & 4B IT & 1200 & 59.91 & 75.60 & 74.32 & 29.38 & 86.98 & 82.00 & 83.30 & 64.16 & 41.19 & 85.14\\
 & 12B IT & 1200 & \underline{70.24} & \underline{86.13} & \underline{84.97} & \underline{21.03} & \underline{88.74} & \textbf{92.40} & \textbf{94.80} & \underline{81.37} & \underline{23.10} & \underline{88.22}\\
\rowcolor{PCTTint}  & \pct{} & 1200 & \textbf{76.36} & \textbf{88.32} & \textbf{87.58} & \textbf{16.02} & \textbf{89.58} & \underline{90.10} & \underline{94.60} & \textbf{85.46} & \textbf{16.95} & \textbf{89.33}\\
\addlinespace[2pt]
$zh\!\to\!en$ & 4B IT & 1213 & 38.89 & 64.46 & 62.03 & 59.53 & 86.20 & 64.40 & 65.90 & 53.20 & 73.92 & 81.00\\
 & 12B IT & 1213 & \underline{47.49} & \underline{75.48} & \underline{72.87} & \underline{40.65} & \underline{87.93} & \textbf{95.70} & \underline{94.90} & \underline{73.06} & \underline{39.17} & \underline{88.17}\\
\rowcolor{PCTTint}  & \pct{} & 1213 & \textbf{58.71} & \textbf{79.80} & \textbf{77.87} & \textbf{32.67} & \textbf{89.16} & \underline{95.20} & \textbf{95.60} & \textbf{78.02} & \textbf{30.94} & \textbf{89.30}\\
\addlinespace[2pt]
$he\!\to\!en$ & 4B IT & 1185 & 63.49 & 77.80 & 76.72 & 28.99 & 87.63 & \textbf{93.40} & \textbf{92.40} & 81.10 & 21.63 & 87.26\\
 & 12B IT & 1185 & \underline{73.72} & \underline{87.81} & \underline{86.82} & \underline{16.72} & \underline{89.22} & \underline{91.80} & 89.20 & \underline{86.07} & \underline{14.94} & \underline{87.78}\\
\rowcolor{PCTTint}  & \pct{} & 1185 & \textbf{79.87} & \textbf{90.07} & \textbf{89.47} & \textbf{12.54} & \textbf{90.03} & 90.80 & \underline{90.00} & \textbf{89.03} & \textbf{10.89} & \textbf{88.69}\\
\addlinespace[2pt]
$fa\!\to\!en$ & 4B IT & 1234 & 57.78 & 76.96 & 75.53 & 29.13 & 88.46 & 95.10 & 95.40 & 77.19 & 27.66 & 87.40\\
 & 12B IT & 1234 & \underline{67.46} & \underline{84.48} & \underline{83.21} & \underline{20.09} & \underline{89.29} & \underline{96.60} & \underline{95.80} & \underline{80.82} & \underline{22.40} & \underline{87.88}\\
\rowcolor{PCTTint}  & \pct{} & 1234 & \textbf{76.31} & \textbf{88.01} & \textbf{87.22} & \textbf{14.41} & \textbf{90.16} & \textbf{96.70} & \textbf{96.70} & \textbf{85.43} & \textbf{16.07} & \textbf{88.65}\\
\addlinespace[2pt]
$es\!\to\!en$ & 4B IT & 1178 & 56.34 & 75.84 & 73.42 & 34.73 & 87.71 & \underline{96.40} & \underline{97.20} & 86.98 & 16.42 & 89.07\\
 & 12B IT & 1178 & \underline{76.88} & \underline{89.96} & \underline{88.91} & \underline{14.43} & \underline{89.31} & \textbf{97.40} & \textbf{97.40} & \underline{88.29} & \underline{13.85} & \underline{89.11}\\
\rowcolor{PCTTint}  & \pct{} & 1178 & \textbf{84.06} & \textbf{92.21} & \textbf{91.63} & \textbf{9.37} & \textbf{90.24} & 94.80 & 96.40 & \textbf{91.34} & \textbf{9.19} & \textbf{89.94}\\
\bottomrule\end{tabular}
\caption{All systems translating into English. Columns through COMET use the full direction-specific test set ($N$). The last 5 columns use the ten shared judge-subset items per direction. Bold/underline mark first/second within the same direction and metric; TER is lower-is-better.}
\end{table*}
\begin{table*}[t]\centering\scriptsize
\setlength{\tabcolsep}{2.8pt}
\renewcommand{\arraystretch}{1.08}
\begin{tabular}{llrrrrrrrrrrr}
\toprule
\rowcolor{TableHeaderDark} {\color{white}\bfseries Direction} & {\color{white}\bfseries System} & {\color{white}\bfseries $N$} & {\color{white}\bfseries BLEU} & {\color{white}\bfseries chrF} & {\color{white}\bfseries chrF++} & {\color{white}\bfseries TER$\downarrow$} & {\color{white}\bfseries COMET} & {\color{white}\bfseries Ref. J.} & {\color{white}\bfseries No-ref. J.} & {\color{white}\bfseries chrF++$_{10}$} & {\color{white}\bfseries TER$_{10}\downarrow$} & {\color{white}\bfseries COMET$_{10}$}\\
\midrule
$ar\!\to\!he$ & 4B IT & 1255 & 33.86 & 55.36 & 53.52 & 55.20 & 85.90 & 70.40 & 66.30 & 48.01 & 52.39 & 84.15\\
 & 12B IT & 1255 & \underline{48.39} & \underline{68.91} & \underline{67.01} & \underline{37.34} & \underline{88.91} & \underline{91.70} & \underline{93.20} & \underline{70.71} & \underline{30.88} & \underline{88.60}\\
\rowcolor{PCTTint}  & \pct{} & 1255 & \textbf{61.41} & \textbf{76.53} & \textbf{75.50} & \textbf{28.87} & \textbf{91.03} & \textbf{93.00} & \textbf{94.70} & \textbf{78.93} & \textbf{20.92} & \textbf{90.49}\\
\addlinespace[2pt]
$zh\!\to\!he$ & 4B IT & 1232 & 19.81 & 39.55 & 38.42 & \underline{88.40} & 82.50 & 85.00 & 87.00 & 53.38 & 61.20 & 85.05\\
 & 12B IT & 1232 & \underline{27.88} & \underline{48.10} & \underline{47.27} & \textbf{78.13} & \underline{85.34} & \underline{90.30} & \underline{88.60} & \underline{58.72} & \underline{53.58} & \underline{86.56}\\
\rowcolor{PCTTint}  & \pct{} & 1232 & \textbf{35.82} & \textbf{51.34} & \textbf{51.36} & 90.09 & \textbf{86.74} & \textbf{93.70} & \textbf{94.70} & \textbf{63.72} & \textbf{46.14} & \textbf{87.24}\\
\addlinespace[2pt]
$en\!\to\!he$ & 4B IT & 1216 & 36.21 & 61.46 & 59.13 & 49.77 & 87.17 & 71.20 & 77.20 & 59.56 & 44.59 & 86.58\\
 & 12B IT & 1216 & \underline{44.79} & \underline{65.92} & \underline{64.17} & \underline{41.62} & \underline{88.31} & \underline{92.60} & \underline{92.10} & \underline{64.07} & \underline{38.82} & \underline{88.78}\\
\rowcolor{PCTTint}  & \pct{} & 1216 & \textbf{56.73} & \textbf{73.34} & \textbf{72.04} & \textbf{32.13} & \textbf{89.85} & \textbf{93.40} & \textbf{92.80} & \textbf{71.85} & \textbf{28.26} & \textbf{89.52}\\
\addlinespace[2pt]
$fa\!\to\!he$ & 4B IT & 1238 & 33.83 & 58.93 & 56.54 & 52.35 & 87.53 & 70.40 & 75.60 & 54.20 & 51.63 & 87.48\\
 & 12B IT & 1238 & \underline{45.84} & \underline{69.08} & \underline{66.79} & \underline{38.09} & \underline{89.74} & \underline{90.10} & \underline{91.20} & \underline{72.00} & \underline{32.67} & \underline{90.98}\\
\rowcolor{PCTTint}  & \pct{} & 1238 & \textbf{58.38} & \textbf{76.09} & \textbf{74.67} & \textbf{33.36} & \textbf{90.94} & \textbf{92.10} & \textbf{94.50} & \textbf{79.33} & \textbf{23.28} & \textbf{92.35}\\
\addlinespace[2pt]
$es\!\to\!he$ & 4B IT & 1255 & 35.99 & 59.59 & 57.50 & 50.63 & 86.26 & 83.00 & 86.10 & 65.78 & 53.33 & 88.65\\
 & 12B IT & 1255 & \underline{49.71} & \underline{70.65} & \underline{68.68} & \underline{35.51} & \underline{88.55} & \underline{90.70} & \underline{91.90} & \underline{73.23} & \underline{29.21} & \underline{89.87}\\
\rowcolor{PCTTint}  & \pct{} & 1255 & \textbf{61.20} & \textbf{77.18} & \textbf{75.92} & \textbf{26.40} & \textbf{89.85} & \textbf{93.60} & \textbf{94.80} & \textbf{80.26} & \textbf{21.41} & \textbf{90.87}\\
\bottomrule\end{tabular}
\caption{All systems translating into Hebrew. Columns through COMET use the full direction-specific test set ($N$). The last 5 columns use the ten shared judge-subset items per direction. Bold/underline mark first/second within the same direction and metric; TER is lower-is-better.}
\end{table*}
\begin{table*}[t]\centering\scriptsize
\setlength{\tabcolsep}{2.8pt}
\renewcommand{\arraystretch}{1.08}
\begin{tabular}{llrrrrrrrrrrr}
\toprule
\rowcolor{TableHeaderDark} {\color{white}\bfseries Direction} & {\color{white}\bfseries System} & {\color{white}\bfseries $N$} & {\color{white}\bfseries BLEU} & {\color{white}\bfseries chrF} & {\color{white}\bfseries chrF++} & {\color{white}\bfseries TER$\downarrow$} & {\color{white}\bfseries COMET} & {\color{white}\bfseries Ref. J.} & {\color{white}\bfseries No-ref. J.} & {\color{white}\bfseries chrF++$_{10}$} & {\color{white}\bfseries TER$_{10}\downarrow$} & {\color{white}\bfseries COMET$_{10}$}\\
\midrule
$ar\!\to\!fa$ & 4B IT & 1228 & 32.96 & 57.46 & 55.29 & 56.93 & 84.60 & 73.20 & 75.10 & 58.50 & 48.14 & 84.77\\
 & 12B IT & 1228 & \underline{45.58} & \underline{71.82} & \underline{69.85} & \underline{40.84} & \underline{88.30} & \textbf{89.80} & \underline{92.60} & \underline{70.16} & \underline{34.38} & \underline{89.77}\\
\rowcolor{PCTTint}  & \pct{} & 1228 & \textbf{62.29} & \textbf{80.41} & \textbf{79.05} & \textbf{23.95} & \textbf{90.46} & \underline{89.50} & \textbf{95.10} & \textbf{80.22} & \textbf{21.59} & \textbf{91.29}\\
\addlinespace[2pt]
$zh\!\to\!fa$ & 4B IT & 1208 & 19.17 & 41.95 & 40.29 & 90.09 & 81.51 & 51.60 & 52.70 & 32.82 & 98.86 & 79.60\\
 & 12B IT & 1208 & \underline{32.94} & \underline{63.68} & \underline{61.27} & \underline{55.33} & \underline{87.85} & \underline{89.40} & \underline{91.50} & \underline{63.25} & \underline{49.32} & \underline{89.01}\\
\rowcolor{PCTTint}  & \pct{} & 1208 & \textbf{45.05} & \textbf{69.11} & \textbf{67.38} & \textbf{55.17} & \textbf{89.04} & \textbf{91.00} & \textbf{93.10} & \textbf{69.82} & \textbf{36.28} & \textbf{90.41}\\
\addlinespace[2pt]
$en\!\to\!fa$ & 4B IT & 1239 & 34.35 & 62.41 & 60.48 & 52.63 & 86.10 & 89.70 & 88.80 & 64.17 & 48.39 & 87.44\\
 & 12B IT & 1239 & \underline{42.64} & \underline{68.48} & \underline{66.60} & \underline{41.52} & \underline{88.24} & \underline{91.70} & \underline{91.90} & \underline{67.09} & \underline{42.60} & \underline{88.90}\\
\rowcolor{PCTTint}  & \pct{} & 1239 & \textbf{56.25} & \textbf{76.24} & \textbf{74.81} & \textbf{29.15} & \textbf{90.13} & \textbf{93.00} & \textbf{93.80} & \textbf{76.46} & \textbf{27.69} & \textbf{90.36}\\
\addlinespace[2pt]
$he\!\to\!fa$ & 4B IT & 1198 & 40.23 & 64.44 & 62.67 & 44.97 & 86.19 & 90.40 & 91.80 & 63.53 & 43.59 & 86.47\\
 & 12B IT & 1198 & \underline{50.15} & \underline{74.71} & \underline{72.89} & \underline{33.55} & \underline{88.98} & \underline{93.40} & \underline{94.80} & \underline{72.82} & \underline{32.88} & \underline{88.59}\\
\rowcolor{PCTTint}  & \pct{} & 1198 & \textbf{63.73} & \textbf{81.50} & \textbf{80.26} & \textbf{23.03} & \textbf{90.18} & \textbf{94.50} & \textbf{95.90} & \textbf{78.81} & \textbf{22.59} & \textbf{89.28}\\
\addlinespace[2pt]
$es\!\to\!fa$ & 4B IT & 1196 & 35.51 & 61.60 & 59.63 & 51.17 & 85.67 & 79.40 & 80.90 & 55.27 & 55.07 & 84.60\\
 & 12B IT & 1196 & \underline{46.74} & \underline{72.49} & \underline{70.47} & \underline{36.71} & \underline{88.22} & \underline{90.90} & \underline{92.20} & \underline{65.24} & \underline{43.73} & \underline{88.41}\\
\rowcolor{PCTTint}  & \pct{} & 1196 & \textbf{59.82} & \textbf{79.44} & \textbf{78.03} & \textbf{27.11} & \textbf{89.71} & \textbf{92.30} & \textbf{93.80} & \textbf{74.01} & \textbf{31.84} & \textbf{90.33}\\
\bottomrule\end{tabular}
\caption{All systems translating into Persian. Columns through COMET use the full direction-specific test set ($N$). The last 5 columns use the ten shared judge-subset items per direction. Bold/underline mark first/second within the same direction and metric; TER is lower-is-better.}
\end{table*}
\begin{table*}[t]\centering\scriptsize
\setlength{\tabcolsep}{2.8pt}
\renewcommand{\arraystretch}{1.08}
\begin{tabular}{llrrrrrrrrrrr}
\toprule
\rowcolor{TableHeaderDark} {\color{white}\bfseries Direction} & {\color{white}\bfseries System} & {\color{white}\bfseries $N$} & {\color{white}\bfseries BLEU} & {\color{white}\bfseries chrF} & {\color{white}\bfseries chrF++} & {\color{white}\bfseries TER$\downarrow$} & {\color{white}\bfseries COMET} & {\color{white}\bfseries Ref. J.} & {\color{white}\bfseries No-ref. J.} & {\color{white}\bfseries chrF++$_{10}$} & {\color{white}\bfseries TER$_{10}\downarrow$} & {\color{white}\bfseries COMET$_{10}$}\\
\midrule
$ar\!\to\!es$ & 4B IT & 1257 & 59.24 & 74.94 & 73.59 & 30.24 & 85.60 & 88.20 & 91.10 & 82.38 & 20.28 & 87.14\\
 & 12B IT & 1257 & \underline{68.19} & \underline{85.36} & \underline{84.08} & \underline{20.85} & \underline{87.93} & \textbf{94.10} & \textbf{96.30} & \underline{86.38} & \underline{14.91} & \underline{88.39}\\
\rowcolor{PCTTint}  & \pct{} & 1257 & \textbf{77.23} & \textbf{88.55} & \textbf{87.73} & \textbf{14.41} & \textbf{89.37} & \underline{91.80} & \underline{94.20} & \textbf{88.63} & \textbf{12.20} & \textbf{88.66}\\
\addlinespace[2pt]
$zh\!\to\!es$ & 4B IT & 1173 & 32.40 & 55.60 & 53.72 & 68.49 & 83.08 & 81.90 & 86.20 & 63.96 & 53.96 & 86.22\\
 & 12B IT & 1173 & \underline{48.63} & \underline{75.81} & \underline{73.37} & \underline{44.50} & \underline{87.27} & \underline{93.60} & \underline{92.90} & \underline{75.41} & \underline{35.64} & \underline{88.58}\\
\rowcolor{PCTTint}  & \pct{} & 1173 & \textbf{58.16} & \textbf{80.03} & \textbf{78.11} & \textbf{31.49} & \textbf{89.03} & \textbf{94.10} & \textbf{94.90} & \textbf{81.32} & \textbf{24.58} & \textbf{89.61}\\
\addlinespace[2pt]
$en\!\to\!es$ & 4B IT & 1188 & 66.60 & 84.25 & 82.63 & 23.34 & 88.08 & 93.20 & 94.90 & 81.53 & 22.75 & 87.87\\
 & 12B IT & 1188 & \underline{73.09} & \underline{87.64} & \underline{86.37} & \underline{17.24} & \underline{88.72} & \textbf{94.30} & \underline{95.90} & \underline{84.14} & \underline{19.54} & \underline{87.99}\\
\rowcolor{PCTTint}  & \pct{} & 1188 & \textbf{79.33} & \textbf{90.12} & \textbf{89.27} & \textbf{12.45} & \textbf{89.61} & \underline{93.90} & \textbf{96.20} & \textbf{86.92} & \textbf{14.98} & \textbf{88.84}\\
\addlinespace[2pt]
$he\!\to\!es$ & 4B IT & 1183 & 58.27 & 71.77 & 70.79 & 30.88 & 85.57 & 78.40 & 77.20 & 58.37 & 49.60 & 82.44\\
 & 12B IT & 1183 & \underline{71.49} & \underline{85.66} & \underline{84.60} & \underline{17.18} & \underline{88.34} & \underline{96.00} & \underline{86.70} & \underline{85.00} & \underline{30.21} & \underline{87.60}\\
\rowcolor{PCTTint}  & \pct{} & 1183 & \textbf{79.39} & \textbf{89.05} & \textbf{88.43} & \textbf{11.91} & \textbf{89.55} & \textbf{96.50} & \textbf{97.00} & \textbf{89.36} & \textbf{11.67} & \textbf{89.92}\\
\addlinespace[2pt]
$fa\!\to\!es$ & 4B IT & 1241 & 59.54 & 76.89 & 75.45 & 29.25 & 87.78 & 83.50 & 89.90 & 77.75 & 27.47 & \underline{86.37}\\
 & 12B IT & 1241 & \underline{67.83} & \underline{84.70} & \underline{83.34} & \underline{20.90} & \underline{89.22} & \underline{94.20} & \underline{92.30} & \underline{79.56} & \underline{24.97} & 86.03\\
\rowcolor{PCTTint}  & \pct{} & 1241 & \textbf{76.09} & \textbf{88.16} & \textbf{87.27} & \textbf{15.38} & \textbf{90.28} & \textbf{94.90} & \textbf{93.30} & \textbf{84.07} & \textbf{20.04} & \textbf{87.36}\\
\bottomrule\end{tabular}
\caption{All systems translating into Spanish. Columns through COMET use the full direction-specific test set ($N$). The last 5 columns use the ten shared judge-subset items per direction. Bold/underline mark first/second within the same direction and metric; TER is lower-is-better.}
\end{table*}
\begin{table*}[t]\centering\scriptsize
\setlength{\tabcolsep}{2.8pt}
\renewcommand{\arraystretch}{1.08}
\begin{tabular}{llrrrrrrrrrrrrrrrr}
\toprule
\rowcolor{TableHeader}  & & \multicolumn{8}{c}{Reference-assisted judge} & \multicolumn{8}{c}{Reference-free judge}\\\cmidrule(lr){3-10}\cmidrule(lr){11-18}
\rowcolor{TableHeaderDark} {\color{white}\bfseries Dir.} & {\color{white}\bfseries System} & {\color{white}\bfseries Sem.} & {\color{white}\bfseries Term.} & {\color{white}\bfseries Flu.} & {\color{white}\bfseries Coh.} & {\color{white}\bfseries Ctx.} & {\color{white}\bfseries Lang.} & {\color{white}\bfseries Style} & {\color{white}\bfseries Fmt.} & {\color{white}\bfseries Sem.} & {\color{white}\bfseries Term.} & {\color{white}\bfseries Flu.} & {\color{white}\bfseries Coh.} & {\color{white}\bfseries Ctx.} & {\color{white}\bfseries Lang.} & {\color{white}\bfseries Style} & {\color{white}\bfseries Fmt.}\\
\midrule
$zh\!\to\!ar$ & 4B IT & 72.0 & 66.5 & 73.8 & 72.1 & 71.8 & 73.9 & 68.8 & 73.5 & 74.1 & 69.3 & 74.0 & 72.7 & 72.7 & 74.8 & 71.5 & 75.8\\
 & 12B IT & \textbf{94.8} & \underline{87.5} & \underline{94.2} & \underline{92.8} & \underline{93.8} & \underline{94.6} & \underline{90.5} & \underline{93.0} & \textbf{95.1} & \underline{92.0} & \underline{91.8} & \underline{92.6} & \underline{92.8} & \underline{93.6} & \underline{91.6} & \underline{94.0}\\
\rowcolor{PCTTint}  & \pct{} & \underline{92.4} & \textbf{89.3} & \textbf{95.2} & \textbf{93.3} & \textbf{93.9} & \textbf{94.8} & \textbf{91.3} & \textbf{96.0} & \underline{95.0} & \textbf{92.1} & \textbf{94.3} & \textbf{93.8} & \textbf{94.5} & \textbf{95.0} & \textbf{92.9} & \textbf{95.0}\\
\addlinespace[2pt]
$en\!\to\!ar$ & 4B IT & \underline{91.0} & \underline{80.2} & 90.2 & 91.0 & \underline{90.1} & \underline{89.2} & \underline{85.5} & 91.5 & 88.0 & 79.7 & 88.5 & 88.6 & 86.6 & \underline{89.0} & 85.0 & 91.5\\
 & 12B IT & 83.3 & 76.0 & \underline{94.3} & \underline{94.5} & 82.8 & 84.5 & 81.3 & \underline{95.0} & \underline{94.0} & \underline{88.5} & \underline{93.4} & \underline{92.2} & \underline{93.2} & 84.0 & \underline{91.2} & \underline{94.5}\\
\rowcolor{PCTTint}  & \pct{} & \textbf{95.7} & \textbf{88.4} & \textbf{97.6} & \textbf{96.1} & \textbf{95.9} & \textbf{97.1} & \textbf{93.2} & \textbf{97.3} & \textbf{96.2} & \textbf{95.0} & \textbf{94.3} & \textbf{94.9} & \textbf{95.3} & \textbf{94.9} & \textbf{92.2} & \textbf{95.5}\\
\addlinespace[2pt]
$he\!\to\!ar$ & 4B IT & 79.1 & 73.8 & 80.3 & 80.3 & 79.5 & 80.1 & 77.0 & 88.0 & 77.0 & 71.2 & 77.4 & 78.6 & 76.1 & 77.0 & 75.5 & 79.8\\
 & 12B IT & \underline{93.8} & \underline{83.2} & \underline{92.4} & \underline{91.2} & \underline{91.6} & \underline{90.3} & \underline{88.2} & \underline{88.5} & \underline{90.7} & \underline{85.5} & \underline{88.9} & \underline{89.5} & \underline{89.7} & \underline{88.1} & \underline{88.7} & \underline{88.5}\\
\rowcolor{PCTTint}  & \pct{} & \textbf{96.3} & \textbf{89.7} & \textbf{96.0} & \textbf{94.7} & \textbf{92.8} & \textbf{95.6} & \textbf{92.0} & \textbf{96.0} & \textbf{95.3} & \textbf{93.7} & \textbf{94.8} & \textbf{94.2} & \textbf{93.9} & \textbf{95.1} & \textbf{91.9} & \textbf{93.9}\\
\addlinespace[2pt]
$fa\!\to\!ar$ & 4B IT & 92.7 & 84.0 & 93.8 & 92.9 & 91.3 & 93.2 & 89.9 & 94.0 & 93.2 & 88.1 & 91.1 & 91.2 & 91.1 & 91.1 & 88.4 & 90.0\\
 & 12B IT & \underline{96.5} & \underline{87.1} & \underline{95.7} & \underline{95.5} & \underline{94.8} & \underline{95.4} & \underline{92.8} & \underline{94.5} & \underline{95.8} & \underline{93.4} & \underline{94.1} & \underline{94.2} & \underline{94.3} & \underline{95.0} & \underline{92.9} & \underline{94.0}\\
\rowcolor{PCTTint}  & \pct{} & \textbf{98.3} & \textbf{91.7} & \textbf{98.2} & \textbf{97.5} & \textbf{96.5} & \textbf{97.9} & \textbf{95.0} & \textbf{97.9} & \textbf{97.0} & \textbf{95.1} & \textbf{95.8} & \textbf{95.3} & \textbf{96.2} & \textbf{95.8} & \textbf{94.2} & \textbf{94.4}\\
\addlinespace[2pt]
$es\!\to\!ar$ & 4B IT & 90.8 & 81.0 & 89.6 & \underline{90.4} & 88.7 & 87.9 & 86.9 & 88.0 & 89.8 & 83.1 & \underline{87.1} & 88.8 & 89.4 & 88.2 & 86.9 & 90.5\\
 & 12B IT & \underline{95.5} & \underline{87.6} & \textbf{96.1} & \textbf{95.3} & \underline{94.7} & \textbf{95.3} & \textbf{92.0} & \underline{93.5} & \textbf{95.7} & \textbf{93.6} & \textbf{92.9} & \underline{92.8} & \textbf{94.5} & \underline{93.8} & \textbf{91.4} & \textbf{94.5}\\
\rowcolor{PCTTint}  & \pct{} & \textbf{95.9} & \textbf{89.8} & \underline{95.8} & \textbf{95.3} & \textbf{94.8} & \underline{95.1} & \underline{91.3} & \textbf{95.0} & \underline{94.2} & \underline{91.0} & \textbf{92.9} & \textbf{93.3} & \underline{92.7} & \textbf{94.0} & \underline{90.5} & \underline{94.3}\\
\bottomrule\end{tabular}
\caption{Every judge dimension for translations into Arabic, ten shared items per direction. Sem.: semantic accuracy; Term.: terminology; Flu.: fluency; Coh.: cohesion; Ctx.: contextual appropriateness; Lang.: target-language adherence; Fmt.: formatting. Scores are model judgments on 0--100 scales, not validated accuracy percentages. Bold/underline mark first/second within each direction and view; ties at displayed precision share rank.}
\end{table*}
\begin{table*}[t]\centering\scriptsize
\setlength{\tabcolsep}{2.8pt}
\renewcommand{\arraystretch}{1.08}
\begin{tabular}{llrrrrrrrrrrrrrrrr}
\toprule
\rowcolor{TableHeader}  & & \multicolumn{8}{c}{Reference-assisted judge} & \multicolumn{8}{c}{Reference-free judge}\\\cmidrule(lr){3-10}\cmidrule(lr){11-18}
\rowcolor{TableHeaderDark} {\color{white}\bfseries Dir.} & {\color{white}\bfseries System} & {\color{white}\bfseries Sem.} & {\color{white}\bfseries Term.} & {\color{white}\bfseries Flu.} & {\color{white}\bfseries Coh.} & {\color{white}\bfseries Ctx.} & {\color{white}\bfseries Lang.} & {\color{white}\bfseries Style} & {\color{white}\bfseries Fmt.} & {\color{white}\bfseries Sem.} & {\color{white}\bfseries Term.} & {\color{white}\bfseries Flu.} & {\color{white}\bfseries Coh.} & {\color{white}\bfseries Ctx.} & {\color{white}\bfseries Lang.} & {\color{white}\bfseries Style} & {\color{white}\bfseries Fmt.}\\
\midrule
$ar\!\to\!zh$ & 4B IT & 75.8 & 72.4 & 74.9 & 73.7 & 74.5 & 75.1 & 72.1 & 74.5 & 76.7 & 72.9 & 75.3 & 74.4 & 74.8 & 75.3 & 72.4 & 74.0\\
 & 12B IT & \underline{95.6} & \underline{88.5} & \underline{94.1} & \underline{93.8} & \underline{94.7} & \underline{93.6} & \underline{90.6} & \underline{91.5} & \underline{95.1} & \underline{91.8} & \underline{94.4} & \underline{92.4} & \underline{93.3} & \underline{92.4} & \underline{90.8} & \underline{91.5}\\
\rowcolor{PCTTint}  & \pct{} & \textbf{95.9} & \textbf{92.1} & \textbf{96.5} & \textbf{95.3} & \textbf{95.6} & \textbf{96.3} & \textbf{92.5} & \textbf{96.4} & \textbf{96.0} & \textbf{94.0} & \textbf{95.4} & \textbf{94.1} & \textbf{94.5} & \textbf{95.7} & \textbf{92.5} & \textbf{95.0}\\
\addlinespace[2pt]
$en\!\to\!zh$ & 4B IT & 84.8 & 80.5 & 80.7 & 81.1 & 83.1 & 82.5 & 80.0 & \underline{92.5} & 86.0 & 81.7 & 90.9 & \underline{93.5} & 84.2 & 82.7 & 81.2 & \underline{94.5}\\
 & 12B IT & \underline{95.4} & \textbf{90.4} & \underline{91.0} & \underline{92.4} & \underline{93.5} & \underline{92.3} & \underline{88.8} & 90.5 & \underline{96.0} & \textbf{93.7} & \underline{93.2} & 93.4 & \underline{94.3} & \underline{93.9} & \underline{92.2} & 93.5\\
\rowcolor{PCTTint}  & \pct{} & \textbf{96.3} & \underline{90.0} & \textbf{95.7} & \textbf{94.1} & \textbf{96.0} & \textbf{96.2} & \textbf{91.7} & \textbf{96.0} & \textbf{96.2} & \underline{92.5} & \textbf{95.2} & \textbf{94.0} & \textbf{94.7} & \textbf{95.5} & \textbf{92.4} & \textbf{95.0}\\
\addlinespace[2pt]
$he\!\to\!zh$ & 4B IT & 80.6 & 69.1 & 80.5 & 80.4 & 80.4 & 82.3 & 77.9 & 81.5 & \underline{89.5} & \underline{79.4} & 92.8 & 91.0 & 90.1 & 82.7 & 89.2 & 93.5\\
 & 12B IT & \underline{96.0} & \textbf{91.9} & \underline{94.4} & \underline{94.3} & \underline{94.9} & \underline{93.7} & \underline{91.2} & \underline{93.0} & \textbf{96.1} & \textbf{94.0} & \underline{93.6} & \underline{93.5} & \textbf{94.4} & \underline{94.1} & \underline{92.9} & \underline{94.7}\\
\rowcolor{PCTTint}  & \pct{} & \textbf{97.1} & \underline{91.3} & \textbf{96.5} & \textbf{95.9} & \textbf{95.8} & \textbf{96.4} & \textbf{94.1} & \textbf{97.0} & \textbf{96.1} & \textbf{94.0} & \textbf{95.1} & \textbf{94.5} & \underline{94.2} & \textbf{95.5} & \textbf{93.4} & \textbf{95.0}\\
\addlinespace[2pt]
$fa\!\to\!zh$ & 4B IT & 76.5 & 70.7 & 75.2 & 75.9 & 75.6 & 75.7 & 72.0 & 75.5 & 85.9 & 82.3 & 85.4 & 85.0 & 75.6 & 75.7 & 74.2 & 84.8\\
 & 12B IT & \underline{94.8} & \underline{90.0} & \underline{94.9} & \underline{93.4} & \underline{93.2} & \underline{95.3} & \underline{90.4} & \underline{94.0} & \underline{93.8} & \underline{87.5} & \underline{93.1} & \underline{92.5} & \underline{93.3} & \underline{94.0} & \underline{90.8} & \underline{92.0}\\
\rowcolor{PCTTint}  & \pct{} & \textbf{97.0} & \textbf{92.8} & \textbf{96.4} & \textbf{95.9} & \textbf{95.3} & \textbf{96.5} & \textbf{94.0} & \textbf{96.3} & \textbf{96.1} & \textbf{93.6} & \textbf{96.0} & \textbf{94.9} & \textbf{95.6} & \textbf{95.9} & \textbf{94.6} & \textbf{94.5}\\
\addlinespace[2pt]
$es\!\to\!zh$ & 4B IT & 72.6 & 63.3 & 69.8 & 71.1 & 70.6 & 68.6 & 66.4 & 65.0 & 74.6 & 68.5 & 72.0 & 72.3 & 72.7 & 72.5 & 70.7 & 70.0\\
 & 12B IT & \underline{82.7} & \underline{79.1} & \underline{82.3} & \underline{82.4} & \underline{82.8} & \underline{82.3} & \underline{79.8} & \underline{83.5} & \underline{86.3} & \underline{85.5} & \underline{85.6} & \underline{85.0} & \underline{85.8} & \underline{89.0} & \underline{84.3} & \underline{84.0}\\
\rowcolor{PCTTint}  & \pct{} & \textbf{86.5} & \textbf{82.6} & \textbf{86.7} & \textbf{84.9} & \textbf{85.7} & \textbf{86.4} & \textbf{83.7} & \textbf{85.5} & \textbf{87.4} & \textbf{86.2} & \textbf{89.8} & \textbf{86.8} & \textbf{87.1} & \textbf{89.4} & \textbf{85.2} & \textbf{88.0}\\
\bottomrule\end{tabular}
\caption{Every judge dimension for translations into Chinese, ten shared items per direction. Sem.: semantic accuracy; Term.: terminology; Flu.: fluency; Coh.: cohesion; Ctx.: contextual appropriateness; Lang.: target-language adherence; Fmt.: formatting. Scores are model judgments on 0--100 scales, not validated accuracy percentages. Bold/underline mark first/second within each direction and view; ties at displayed precision share rank.}
\end{table*}
\begin{table*}[t]\centering\scriptsize
\setlength{\tabcolsep}{2.8pt}
\renewcommand{\arraystretch}{1.08}
\begin{tabular}{llrrrrrrrrrrrrrrrr}
\toprule
\rowcolor{TableHeader}  & & \multicolumn{8}{c}{Reference-assisted judge} & \multicolumn{8}{c}{Reference-free judge}\\\cmidrule(lr){3-10}\cmidrule(lr){11-18}
\rowcolor{TableHeaderDark} {\color{white}\bfseries Dir.} & {\color{white}\bfseries System} & {\color{white}\bfseries Sem.} & {\color{white}\bfseries Term.} & {\color{white}\bfseries Flu.} & {\color{white}\bfseries Coh.} & {\color{white}\bfseries Ctx.} & {\color{white}\bfseries Lang.} & {\color{white}\bfseries Style} & {\color{white}\bfseries Fmt.} & {\color{white}\bfseries Sem.} & {\color{white}\bfseries Term.} & {\color{white}\bfseries Flu.} & {\color{white}\bfseries Coh.} & {\color{white}\bfseries Ctx.} & {\color{white}\bfseries Lang.} & {\color{white}\bfseries Style} & {\color{white}\bfseries Fmt.}\\
\midrule
$ar\!\to\!en$ & 4B IT & 83.3 & 75.0 & 85.0 & 84.0 & 83.7 & 85.1 & 80.6 & 86.5 & 86.2 & 82.8 & 84.4 & 84.3 & \underline{83.7} & 85.2 & 81.3 & 85.5\\
 & 12B IT & \textbf{95.1} & \textbf{89.0} & \underline{95.5} & \underline{94.5} & \underline{94.5} & \underline{95.5} & \underline{91.2} & \underline{93.0} & \underline{96.3} & \underline{94.4} & \textbf{95.5} & \textbf{95.3} & \textbf{95.4} & \underline{95.7} & \textbf{93.7} & \underline{94.5}\\
\rowcolor{PCTTint}  & \pct{} & \underline{91.2} & \underline{87.4} & \textbf{96.9} & \textbf{96.9} & \textbf{96.3} & \textbf{97.1} & \textbf{94.1} & \textbf{97.0} & \textbf{96.6} & \textbf{94.7} & \underline{95.2} & \underline{94.5} & \textbf{95.4} & \textbf{96.2} & \underline{93.5} & \textbf{95.1}\\
\addlinespace[2pt]
$zh\!\to\!en$ & 4B IT & 65.1 & 61.5 & 64.1 & 65.8 & \underline{64.9} & 63.8 & 62.0 & 75.5 & \underline{68.0} & \underline{65.0} & 75.0 & 75.4 & 65.8 & 66.6 & 65.3 & 76.0\\
 & 12B IT & \textbf{97.9} & \textbf{95.3} & \underline{95.9} & \textbf{96.9} & \textbf{96.7} & \underline{96.1} & \textbf{94.4} & \underline{95.2} & \textbf{97.1} & \textbf{96.4} & \underline{94.8} & \underline{94.8} & \underline{95.0} & \underline{96.2} & \underline{93.1} & \underline{94.5}\\
\rowcolor{PCTTint}  & \pct{} & \underline{97.5} & \underline{93.6} & \textbf{96.8} & \underline{96.4} & \textbf{96.7} & \textbf{96.7} & \underline{94.1} & \textbf{96.0} & \textbf{97.1} & \textbf{96.4} & \textbf{96.8} & \textbf{95.4} & \textbf{95.9} & \textbf{97.1} & \textbf{94.6} & \textbf{95.8}\\
\addlinespace[2pt]
$he\!\to\!en$ & 4B IT & \textbf{95.6} & \textbf{88.8} & 96.4 & 95.7 & \textbf{95.6} & \underline{97.1} & 92.6 & 95.0 & \textbf{93.2} & \textbf{90.7} & \underline{95.8} & \textbf{95.0} & \textbf{94.5} & \textbf{96.2} & \underline{93.6} & 94.0\\
 & 12B IT & \underline{92.4} & 86.3 & \underline{97.1} & \underline{97.2} & \underline{95.1} & 96.8 & \underline{93.5} & \underline{96.0} & 90.3 & 86.7 & \textbf{95.9} & 94.7 & 90.5 & 95.2 & 93.2 & \underline{94.3}\\
\rowcolor{PCTTint}  & \pct{} & 91.4 & \underline{88.4} & \textbf{98.4} & \textbf{98.2} & 92.7 & \textbf{98.9} & \textbf{95.5} & \textbf{98.5} & \underline{90.9} & \underline{89.1} & 94.9 & \underline{94.8} & \underline{91.1} & \underline{95.9} & \textbf{93.7} & \textbf{96.3}\\
\addlinespace[2pt]
$fa\!\to\!en$ & 4B IT & 97.5 & 94.0 & 96.9 & 97.0 & 97.3 & 97.0 & \underline{94.1} & 94.0 & 96.5 & 95.8 & 94.7 & 95.9 & 96.2 & 95.5 & \underline{95.0} & \underline{95.0}\\
 & 12B IT & \textbf{98.4} & \textbf{97.5} & \underline{97.6} & \underline{97.7} & \underline{98.7} & \underline{97.7} & \textbf{95.1} & \underline{95.8} & \underline{97.8} & \underline{96.4} & \underline{95.6} & \underline{96.0} & \underline{96.9} & \underline{96.7} & 94.7 & 94.5\\
\rowcolor{PCTTint}  & \pct{} & \underline{97.9} & \underline{97.3} & \textbf{99.2} & \textbf{97.9} & \textbf{98.9} & \textbf{99.3} & \textbf{95.1} & \textbf{98.0} & \textbf{98.2} & \textbf{97.7} & \textbf{96.7} & \textbf{96.7} & \textbf{97.4} & \textbf{97.4} & \textbf{96.1} & \textbf{95.7}\\
\addlinespace[2pt]
$es\!\to\!en$ & 4B IT & \underline{99.1} & \underline{96.3} & \textbf{97.8} & \underline{98.3} & \underline{98.4} & \textbf{98.2} & \underline{95.1} & 96.7 & \underline{98.3} & \underline{98.6} & \underline{97.3} & 97.3 & \underline{97.9} & \underline{97.5} & \underline{96.6} & 96.0\\
 & 12B IT & \textbf{99.7} & \textbf{98.9} & 97.0 & \textbf{99.3} & \textbf{99.6} & \underline{97.8} & \textbf{95.5} & \underline{97.9} & \textbf{99.0} & \textbf{99.0} & \underline{97.3} & \underline{97.6} & \textbf{98.7} & 97.3 & \textbf{96.8} & \underline{96.3}\\
\rowcolor{PCTTint}  & \pct{} & 96.2 & 95.0 & \underline{97.5} & 97.4 & 96.6 & 97.6 & 93.6 & \textbf{98.0} & 97.3 & 96.4 & \textbf{97.5} & \textbf{97.7} & 97.1 & \textbf{97.6} & \underline{96.6} & \textbf{97.5}\\
\bottomrule\end{tabular}
\caption{Every judge dimension for translations into English, ten shared items per direction. Sem.: semantic accuracy; Term.: terminology; Flu.: fluency; Coh.: cohesion; Ctx.: contextual appropriateness; Lang.: target-language adherence; Fmt.: formatting. Scores are model judgments on 0--100 scales, not validated accuracy percentages. Bold/underline mark first/second within each direction and view; ties at displayed precision share rank.}
\end{table*}
\begin{table*}[t]\centering\scriptsize
\setlength{\tabcolsep}{2.8pt}
\renewcommand{\arraystretch}{1.08}
\begin{tabular}{llrrrrrrrrrrrrrrrr}
\toprule
\rowcolor{TableHeader}  & & \multicolumn{8}{c}{Reference-assisted judge} & \multicolumn{8}{c}{Reference-free judge}\\\cmidrule(lr){3-10}\cmidrule(lr){11-18}
\rowcolor{TableHeaderDark} {\color{white}\bfseries Dir.} & {\color{white}\bfseries System} & {\color{white}\bfseries Sem.} & {\color{white}\bfseries Term.} & {\color{white}\bfseries Flu.} & {\color{white}\bfseries Coh.} & {\color{white}\bfseries Ctx.} & {\color{white}\bfseries Lang.} & {\color{white}\bfseries Style} & {\color{white}\bfseries Fmt.} & {\color{white}\bfseries Sem.} & {\color{white}\bfseries Term.} & {\color{white}\bfseries Flu.} & {\color{white}\bfseries Coh.} & {\color{white}\bfseries Ctx.} & {\color{white}\bfseries Lang.} & {\color{white}\bfseries Style} & {\color{white}\bfseries Fmt.}\\
\midrule
$ar\!\to\!he$ & 4B IT & 72.5 & 64.0 & 72.0 & 72.5 & 72.0 & 72.0 & 68.5 & 83.0 & 69.4 & 62.5 & 69.5 & 69.9 & 69.2 & 70.0 & 66.6 & 69.5\\
 & 12B IT & \underline{96.0} & \underline{86.3} & \underline{94.5} & \underline{94.4} & \underline{95.0} & \underline{95.0} & \underline{90.0} & \underline{94.5} & \underline{96.3} & \underline{92.9} & \underline{95.2} & \underline{93.9} & \underline{94.1} & \underline{94.5} & \underline{92.3} & \underline{92.1}\\
\rowcolor{PCTTint}  & \pct{} & \textbf{96.3} & \textbf{88.5} & \textbf{95.6} & \textbf{96.2} & \textbf{96.2} & \textbf{95.6} & \textbf{93.6} & \textbf{97.5} & \textbf{97.6} & \textbf{94.8} & \textbf{95.8} & \textbf{94.9} & \textbf{96.5} & \textbf{95.9} & \textbf{93.8} & \textbf{96.5}\\
\addlinespace[2pt]
$zh\!\to\!he$ & 4B IT & 89.0 & 80.8 & 87.7 & 88.7 & 88.7 & 87.4 & 86.0 & 88.0 & 90.4 & 84.8 & 87.5 & 87.9 & \underline{89.0} & 89.1 & 86.5 & 90.0\\
 & 12B IT & \underline{93.3} & \underline{86.8} & \underline{92.9} & \underline{91.8} & \underline{92.3} & \underline{92.9} & \underline{90.1} & \underline{91.2} & \underline{92.7} & \underline{87.2} & \underline{89.0} & \underline{91.2} & 88.7 & \underline{89.5} & \underline{86.9} & \underline{91.0}\\
\rowcolor{PCTTint}  & \pct{} & \textbf{97.4} & \textbf{90.5} & \textbf{95.9} & \textbf{95.6} & \textbf{96.0} & \textbf{96.5} & \textbf{93.3} & \textbf{97.3} & \textbf{96.4} & \textbf{94.8} & \textbf{95.0} & \textbf{95.3} & \textbf{95.1} & \textbf{95.6} & \textbf{93.2} & \textbf{96.0}\\
\addlinespace[2pt]
$en\!\to\!he$ & 4B IT & 72.0 & 62.5 & 79.5 & 80.0 & 74.5 & 78.5 & 72.5 & 85.5 & 82.2 & 69.5 & 79.5 & 81.7 & 80.0 & 80.0 & 78.5 & 87.0\\
 & 12B IT & \underline{96.6} & \underline{88.1} & \underline{94.8} & \underline{94.8} & \underline{95.9} & \underline{93.8} & \underline{92.1} & \underline{93.0} & \underline{95.0} & \underline{91.1} & \underline{91.6} & \textbf{93.2} & \underline{93.3} & \underline{92.4} & \underline{90.9} & \underline{93.5}\\
\rowcolor{PCTTint}  & \pct{} & \textbf{96.9} & \textbf{91.2} & \textbf{95.6} & \textbf{95.7} & \textbf{96.6} & \textbf{95.5} & \textbf{92.4} & \textbf{97.0} & \textbf{95.6} & \textbf{92.5} & \textbf{92.8} & \underline{93.1} & \textbf{93.9} & \textbf{93.8} & \textbf{91.8} & \textbf{95.5}\\
\addlinespace[2pt]
$fa\!\to\!he$ & 4B IT & 72.0 & 61.5 & 76.3 & 77.5 & 71.5 & 76.3 & 72.5 & 90.5 & 79.0 & 70.8 & 78.6 & 79.2 & 78.1 & 80.5 & 76.1 & 82.0\\
 & 12B IT & \underline{95.1} & \underline{84.0} & \underline{94.6} & \underline{94.2} & \underline{92.3} & \underline{94.1} & \textbf{91.3} & \textbf{95.5} & \underline{94.7} & \underline{90.2} & \underline{92.4} & \underline{91.7} & \underline{92.6} & \underline{92.5} & \underline{90.8} & \underline{94.0}\\
\rowcolor{PCTTint}  & \pct{} & \textbf{95.4} & \textbf{89.0} & \textbf{95.3} & \textbf{94.4} & \textbf{94.9} & \textbf{95.5} & \underline{91.1} & \underline{95.1} & \textbf{96.6} & \textbf{94.9} & \textbf{94.6} & \textbf{94.7} & \textbf{95.9} & \textbf{95.7} & \textbf{92.5} & \textbf{95.6}\\
\addlinespace[2pt]
$es\!\to\!he$ & 4B IT & 88.3 & 78.7 & 85.7 & 87.6 & 86.1 & 84.7 & 83.5 & 93.0 & 88.7 & 82.2 & 87.2 & 88.8 & 88.2 & 87.6 & 86.7 & \underline{93.0}\\
 & 12B IT & \underline{94.0} & \underline{86.5} & \underline{94.2} & \underline{93.1} & \underline{93.8} & \underline{94.2} & \underline{90.4} & \underline{94.0} & \underline{94.2} & \underline{90.0} & \underline{93.4} & \underline{92.6} & \underline{92.3} & \underline{93.1} & \underline{90.8} & \underline{93.0}\\
\rowcolor{PCTTint}  & \pct{} & \textbf{97.2} & \textbf{92.2} & \textbf{96.0} & \textbf{95.2} & \textbf{95.1} & \textbf{95.7} & \textbf{93.0} & \textbf{97.5} & \textbf{97.3} & \textbf{95.3} & \textbf{95.3} & \textbf{95.2} & \textbf{95.8} & \textbf{95.5} & \textbf{93.6} & \textbf{96.0}\\
\bottomrule\end{tabular}
\caption{Every judge dimension for translations into Hebrew, ten shared items per direction. Sem.: semantic accuracy; Term.: terminology; Flu.: fluency; Coh.: cohesion; Ctx.: contextual appropriateness; Lang.: target-language adherence; Fmt.: formatting. Scores are model judgments on 0--100 scales, not validated accuracy percentages. Bold/underline mark first/second within each direction and view; ties at displayed precision share rank.}
\end{table*}
\begin{table*}[t]\centering\scriptsize
\setlength{\tabcolsep}{2.8pt}
\renewcommand{\arraystretch}{1.08}
\begin{tabular}{llrrrrrrrrrrrrrrrr}
\toprule
\rowcolor{TableHeader}  & & \multicolumn{8}{c}{Reference-assisted judge} & \multicolumn{8}{c}{Reference-free judge}\\\cmidrule(lr){3-10}\cmidrule(lr){11-18}
\rowcolor{TableHeaderDark} {\color{white}\bfseries Dir.} & {\color{white}\bfseries System} & {\color{white}\bfseries Sem.} & {\color{white}\bfseries Term.} & {\color{white}\bfseries Flu.} & {\color{white}\bfseries Coh.} & {\color{white}\bfseries Ctx.} & {\color{white}\bfseries Lang.} & {\color{white}\bfseries Style} & {\color{white}\bfseries Fmt.} & {\color{white}\bfseries Sem.} & {\color{white}\bfseries Term.} & {\color{white}\bfseries Flu.} & {\color{white}\bfseries Coh.} & {\color{white}\bfseries Ctx.} & {\color{white}\bfseries Lang.} & {\color{white}\bfseries Style} & {\color{white}\bfseries Fmt.}\\
\midrule
$ar\!\to\!fa$ & 4B IT & 77.0 & 68.8 & 73.4 & 74.8 & 75.5 & 75.3 & 71.0 & 78.0 & 77.6 & 74.1 & 76.0 & 76.3 & 76.2 & 76.6 & 74.3 & 75.5\\
 & 12B IT & \textbf{93.2} & \underline{85.4} & \underline{94.1} & \underline{93.0} & \underline{91.1} & \underline{94.1} & \underline{89.3} & \underline{94.5} & \underline{95.4} & \underline{91.0} & \underline{95.0} & \underline{93.5} & \underline{94.1} & \underline{94.8} & \underline{91.8} & \underline{94.0}\\
\rowcolor{PCTTint}  & \pct{} & \underline{92.2} & \textbf{85.7} & \textbf{96.4} & \textbf{94.5} & \textbf{94.6} & \textbf{96.4} & \textbf{92.0} & \textbf{97.7} & \textbf{97.0} & \textbf{95.2} & \textbf{96.5} & \textbf{96.4} & \textbf{96.8} & \textbf{96.1} & \textbf{95.3} & \textbf{96.5}\\
\addlinespace[2pt]
$zh\!\to\!fa$ & 4B IT & 51.5 & 45.0 & 53.0 & 51.0 & 51.5 & 51.5 & 49.0 & 64.5 & 55.0 & 49.5 & 55.0 & 55.0 & 55.5 & 54.0 & 53.0 & 56.5\\
 & 12B IT & \underline{91.4} & \underline{84.8} & \underline{93.3} & \underline{92.4} & \underline{92.3} & \underline{93.0} & \underline{90.0} & \underline{93.0} & \underline{94.3} & \underline{89.8} & \underline{92.3} & \underline{93.1} & \underline{93.6} & \underline{93.6} & \underline{90.8} & \underline{92.5}\\
\rowcolor{PCTTint}  & \pct{} & \textbf{93.6} & \textbf{86.4} & \textbf{95.1} & \textbf{93.1} & \textbf{94.3} & \textbf{95.4} & \textbf{92.9} & \textbf{94.5} & \textbf{95.2} & \textbf{92.6} & \textbf{94.5} & \textbf{93.8} & \textbf{94.6} & \textbf{94.9} & \textbf{92.8} & \textbf{96.0}\\
\addlinespace[2pt]
$en\!\to\!fa$ & 4B IT & 94.1 & 85.7 & 92.6 & 92.9 & 91.0 & 91.4 & 88.4 & 92.0 & 93.2 & 87.2 & 88.7 & 91.6 & 90.1 & 89.4 & 87.3 & 91.5\\
 & 12B IT & \underline{95.4} & \underline{88.0} & \underline{93.4} & \underline{93.8} & \underline{94.0} & \underline{93.4} & \underline{90.7} & \underline{92.5} & \underline{95.1} & \underline{92.3} & \underline{91.5} & \underline{92.6} & \underline{92.7} & \underline{92.8} & \underline{89.4} & \underline{93.3}\\
\rowcolor{PCTTint}  & \pct{} & \textbf{96.7} & \textbf{90.3} & \textbf{94.4} & \textbf{95.2} & \textbf{95.2} & \textbf{94.3} & \textbf{91.8} & \textbf{93.5} & \textbf{96.3} & \textbf{94.5} & \textbf{93.4} & \textbf{94.0} & \textbf{94.1} & \textbf{94.1} & \textbf{92.1} & \textbf{94.0}\\
\addlinespace[2pt]
$he\!\to\!fa$ & 4B IT & 94.8 & 84.9 & 92.8 & 93.3 & 94.5 & 93.1 & 90.2 & 93.5 & 94.7 & 90.9 & 93.1 & 93.4 & 93.4 & 93.1 & 91.4 & 93.5\\
 & 12B IT & \underline{97.0} & \underline{89.4} & \underline{95.5} & \underline{95.8} & \underline{96.1} & \underline{95.6} & \underline{93.0} & \underline{94.5} & \underline{96.5} & \underline{94.3} & \underline{94.6} & \underline{95.5} & \underline{96.0} & \underline{94.8} & \underline{94.4} & \underline{94.5}\\
\rowcolor{PCTTint}  & \pct{} & \textbf{97.7} & \textbf{90.9} & \textbf{97.5} & \textbf{96.9} & \textbf{96.7} & \textbf{97.3} & \textbf{94.3} & \textbf{96.9} & \textbf{98.0} & \textbf{96.9} & \textbf{96.3} & \textbf{95.7} & \textbf{96.8} & \textbf{96.5} & \textbf{94.5} & \textbf{95.0}\\
\addlinespace[2pt]
$es\!\to\!fa$ & 4B IT & 82.1 & 74.4 & 80.2 & 81.2 & 81.4 & 80.7 & 80.0 & 80.5 & 84.8 & 79.4 & 81.9 & 83.6 & 82.7 & 81.9 & 80.4 & 83.0\\
 & 12B IT & \underline{95.5} & \underline{86.7} & \underline{93.6} & \underline{93.7} & \underline{93.9} & \underline{92.7} & \underline{90.8} & \underline{91.0} & \underline{95.4} & \underline{91.8} & \underline{91.8} & \textbf{93.5} & \underline{94.1} & \underline{92.1} & \underline{91.2} & \underline{94.0}\\
\rowcolor{PCTTint}  & \pct{} & \textbf{96.5} & \textbf{88.2} & \textbf{94.3} & \textbf{95.4} & \textbf{95.6} & \textbf{94.7} & \textbf{92.9} & \textbf{96.0} & \textbf{96.2} & \textbf{94.4} & \textbf{93.5} & \underline{93.4} & \textbf{94.8} & \textbf{94.7} & \textbf{92.2} & \textbf{95.0}\\
\bottomrule\end{tabular}
\caption{Every judge dimension for translations into Persian, ten shared items per direction. Sem.: semantic accuracy; Term.: terminology; Flu.: fluency; Coh.: cohesion; Ctx.: contextual appropriateness; Lang.: target-language adherence; Fmt.: formatting. Scores are model judgments on 0--100 scales, not validated accuracy percentages. Bold/underline mark first/second within each direction and view; ties at displayed precision share rank.}
\end{table*}
\begin{table*}[t]\centering\scriptsize
\setlength{\tabcolsep}{2.8pt}
\renewcommand{\arraystretch}{1.08}
\begin{tabular}{llrrrrrrrrrrrrrrrr}
\toprule
\rowcolor{TableHeader}  & & \multicolumn{8}{c}{Reference-assisted judge} & \multicolumn{8}{c}{Reference-free judge}\\\cmidrule(lr){3-10}\cmidrule(lr){11-18}
\rowcolor{TableHeaderDark} {\color{white}\bfseries Dir.} & {\color{white}\bfseries System} & {\color{white}\bfseries Sem.} & {\color{white}\bfseries Term.} & {\color{white}\bfseries Flu.} & {\color{white}\bfseries Coh.} & {\color{white}\bfseries Ctx.} & {\color{white}\bfseries Lang.} & {\color{white}\bfseries Style} & {\color{white}\bfseries Fmt.} & {\color{white}\bfseries Sem.} & {\color{white}\bfseries Term.} & {\color{white}\bfseries Flu.} & {\color{white}\bfseries Coh.} & {\color{white}\bfseries Ctx.} & {\color{white}\bfseries Lang.} & {\color{white}\bfseries Style} & {\color{white}\bfseries Fmt.}\\
\midrule
$ar\!\to\!es$ & 4B IT & 91.8 & 80.7 & 94.1 & 92.2 & 92.3 & 93.0 & 89.1 & 92.5 & 94.4 & 87.8 & 94.3 & 94.0 & 93.4 & 93.5 & 92.5 & 95.8\\
 & 12B IT & \textbf{97.4} & \textbf{89.5} & \textbf{98.0} & \textbf{97.1} & \textbf{96.9} & \textbf{97.3} & \textbf{93.9} & \underline{93.0} & \textbf{98.5} & \textbf{96.5} & \underline{96.6} & \textbf{96.7} & \textbf{97.5} & \underline{96.6} & \textbf{95.9} & \underline{96.0}\\
\rowcolor{PCTTint}  & \pct{} & \underline{95.0} & \underline{85.5} & \underline{96.8} & \underline{95.8} & \underline{96.1} & \underline{96.1} & \underline{93.8} & \textbf{96.8} & \underline{96.2} & \underline{91.6} & \textbf{97.1} & \underline{95.7} & \underline{95.3} & \textbf{97.1} & \underline{94.3} & \textbf{96.1}\\
\addlinespace[2pt]
$zh\!\to\!es$ & 4B IT & 86.1 & 78.0 & 85.0 & 84.7 & 85.4 & 84.9 & 81.7 & 83.5 & 87.6 & 87.3 & 85.6 & 86.4 & 86.7 & 86.4 & 84.6 & 85.5\\
 & 12B IT & \underline{96.4} & \underline{90.6} & \underline{95.7} & \underline{95.1} & \underline{95.5} & \underline{95.5} & \underline{92.0} & \underline{93.5} & \underline{95.8} & \underline{93.3} & \underline{92.9} & \underline{93.6} & \underline{93.2} & \underline{93.3} & \underline{91.5} & \underline{93.6}\\
\rowcolor{PCTTint}  & \pct{} & \textbf{97.5} & \textbf{91.4} & \textbf{96.6} & \textbf{95.2} & \textbf{96.8} & \textbf{96.3} & \textbf{94.3} & \textbf{96.5} & \textbf{96.5} & \textbf{95.7} & \textbf{94.8} & \textbf{95.1} & \textbf{95.1} & \textbf{95.9} & \textbf{94.0} & \textbf{94.8}\\
\addlinespace[2pt]
$en\!\to\!es$ & 4B IT & 96.9 & 89.8 & 95.5 & 95.8 & 95.7 & 94.6 & 92.9 & 93.5 & 96.9 & 95.6 & 95.3 & 95.3 & \underline{96.6} & 95.8 & 94.4 & 95.0\\
 & 12B IT & \textbf{98.2} & \underline{90.7} & \textbf{97.3} & \underline{97.2} & \textbf{97.2} & \underline{96.4} & \textbf{94.6} & \underline{94.5} & \textbf{98.3} & \textbf{97.3} & \underline{95.5} & \textbf{97.0} & \textbf{97.5} & \underline{96.2} & \underline{95.2} & \textbf{97.5}\\
\rowcolor{PCTTint}  & \pct{} & \underline{97.3} & \textbf{91.6} & \underline{96.9} & \textbf{97.3} & \underline{97.1} & \textbf{96.8} & \underline{94.4} & \textbf{97.5} & \underline{97.5} & \underline{96.4} & \textbf{96.9} & \underline{96.2} & \underline{96.6} & \textbf{97.0} & \textbf{95.6} & \underline{96.2}\\
\addlinespace[2pt]
$he\!\to\!es$ & 4B IT & 80.6 & 75.0 & 79.7 & 80.1 & 79.6 & 79.3 & 76.9 & 95.5 & 78.2 & 76.8 & 77.7 & 77.7 & 77.3 & 78.2 & 75.9 & 80.5\\
 & 12B IT & \textbf{99.2} & \underline{93.6} & \textbf{98.9} & \textbf{98.5} & \underline{98.6} & \underline{98.3} & \textbf{95.7} & \underline{95.7} & \underline{87.8} & \underline{87.2} & \underline{86.8} & \underline{86.8} & \underline{87.5} & \underline{86.7} & \underline{86.1} & \underline{85.5}\\
\rowcolor{PCTTint}  & \pct{} & \underline{99.1} & \textbf{97.5} & \underline{98.3} & \underline{98.2} & \textbf{99.0} & \textbf{98.9} & \underline{94.6} & \textbf{99.0} & \textbf{98.1} & \textbf{98.0} & \textbf{98.2} & \textbf{97.6} & \textbf{98.1} & \textbf{98.4} & \textbf{95.8} & \textbf{95.8}\\
\addlinespace[2pt]
$fa\!\to\!es$ & 4B IT & 84.6 & 81.2 & 94.5 & 93.2 & 88.7 & 93.9 & 87.3 & 92.5 & 92.5 & 90.9 & \underline{94.8} & 92.9 & \underline{92.4} & 94.9 & \textbf{91.6} & \underline{91.0}\\
 & 12B IT & \underline{97.5} & \underline{94.0} & \underline{96.3} & \underline{96.8} & \underline{95.5} & \underline{96.5} & \underline{92.5} & \underline{94.0} & \underline{95.0} & \underline{92.6} & \textbf{95.3} & \underline{93.3} & \underline{92.4} & \textbf{95.5} & \underline{91.5} & \underline{91.0}\\
\rowcolor{PCTTint}  & \pct{} & \textbf{98.1} & \textbf{94.2} & \textbf{96.4} & \textbf{97.0} & \textbf{97.1} & \textbf{97.7} & \textbf{93.3} & \textbf{96.1} & \textbf{96.2} & \textbf{93.8} & 94.0 & \textbf{94.1} & \textbf{93.1} & \underline{95.1} & \textbf{91.6} & \textbf{94.5}\\
\bottomrule\end{tabular}
\caption{Every judge dimension for translations into Spanish, ten shared items per direction. Sem.: semantic accuracy; Term.: terminology; Flu.: fluency; Coh.: cohesion; Ctx.: contextual appropriateness; Lang.: target-language adherence; Fmt.: formatting. Scores are model judgments on 0--100 scales, not validated accuracy percentages. Bold/underline mark first/second within each direction and view; ties at displayed precision share rank.}
\end{table*}
\begin{table*}[t]\centering\scriptsize
\setlength{\tabcolsep}{2.8pt}
\renewcommand{\arraystretch}{1.08}
\begin{tabular}{llrrrrrrrrrrrrrrrr}
\toprule
\rowcolor{TableHeader}  & & \multicolumn{8}{c}{Reference-assisted judge} & \multicolumn{8}{c}{Reference-free judge}\\\cmidrule(lr){3-10}\cmidrule(lr){11-18}
\rowcolor{TableHeaderDark} {\color{white}\bfseries Target} & {\color{white}\bfseries System} & {\color{white}\bfseries Sem.} & {\color{white}\bfseries Term.} & {\color{white}\bfseries Flu.} & {\color{white}\bfseries Coh.} & {\color{white}\bfseries Ctx.} & {\color{white}\bfseries Lang.} & {\color{white}\bfseries Style} & {\color{white}\bfseries Fmt.} & {\color{white}\bfseries Sem.} & {\color{white}\bfseries Term.} & {\color{white}\bfseries Flu.} & {\color{white}\bfseries Coh.} & {\color{white}\bfseries Ctx.} & {\color{white}\bfseries Lang.} & {\color{white}\bfseries Style} & {\color{white}\bfseries Fmt.}\\
\midrule
ar & 4B IT & 85.1 & 77.1 & 85.5 & 85.3 & 84.3 & 84.9 & 81.6 & 87.0 & 84.4 & 78.3 & 83.6 & 84.0 & 83.2 & 84.0 & 81.5 & 85.5\\
 & 12B IT & \underline{92.8} & \underline{84.3} & \underline{94.5} & \underline{93.9} & \underline{91.5} & \underline{92.0} & \underline{89.0} & \underline{92.9} & \underline{94.3} & \underline{90.6} & \underline{92.2} & \underline{92.3} & \underline{92.9} & \underline{90.9} & \underline{91.2} & \underline{93.1}\\
\rowcolor{PCTTint}  & \pct{} & \textbf{95.7} & \textbf{89.8} & \textbf{96.6} & \textbf{95.4} & \textbf{94.8} & \textbf{96.1} & \textbf{92.6} & \textbf{96.4} & \textbf{95.5} & \textbf{93.4} & \textbf{94.4} & \textbf{94.3} & \textbf{94.5} & \textbf{95.0} & \textbf{92.3} & \textbf{94.6}\\
\addlinespace[2pt]
zh & 4B IT & 78.1 & 71.2 & 76.2 & 76.4 & 76.8 & 76.8 & 73.7 & 77.8 & 82.5 & 77.0 & 83.3 & 83.2 & 79.5 & 77.8 & 77.5 & 83.4\\
 & 12B IT & \underline{92.9} & \underline{88.0} & \underline{91.3} & \underline{91.3} & \underline{91.8} & \underline{91.4} & \underline{88.2} & \underline{90.5} & \underline{93.5} & \underline{90.5} & \underline{92.0} & \underline{91.4} & \underline{92.2} & \underline{92.7} & \underline{90.2} & \underline{91.1}\\
\rowcolor{PCTTint}  & \pct{} & \textbf{94.6} & \textbf{89.8} & \textbf{94.4} & \textbf{93.2} & \textbf{93.7} & \textbf{94.4} & \textbf{91.2} & \textbf{94.2} & \textbf{94.4} & \textbf{92.1} & \textbf{94.3} & \textbf{92.9} & \textbf{93.2} & \textbf{94.4} & \textbf{91.6} & \textbf{93.5}\\
\addlinespace[2pt]
en & 4B IT & 88.1 & 83.1 & 88.0 & 88.2 & 88.0 & 88.2 & 84.9 & 89.5 & 88.4 & 86.6 & 89.4 & 89.6 & 87.6 & 88.2 & 86.4 & 89.3\\
 & 12B IT & \textbf{96.7} & \textbf{93.4} & \underline{96.6} & \underline{97.1} & \textbf{96.9} & \underline{96.8} & \underline{93.9} & \underline{95.6} & \textbf{96.1} & \underline{94.6} & \underline{95.8} & \underline{95.7} & \underline{95.3} & \underline{96.2} & \underline{94.3} & \underline{94.8}\\
\rowcolor{PCTTint}  & \pct{} & \underline{94.8} & \underline{92.3} & \textbf{97.8} & \textbf{97.4} & \underline{96.2} & \textbf{97.9} & \textbf{94.5} & \textbf{97.5} & \underline{96.0} & \textbf{94.9} & \textbf{96.2} & \textbf{95.8} & \textbf{95.4} & \textbf{96.8} & \textbf{94.9} & \textbf{96.1}\\
\addlinespace[2pt]
he & 4B IT & 78.8 & 69.5 & 80.2 & 81.3 & 78.6 & 79.8 & 76.6 & 88.0 & 81.9 & 74.0 & 80.5 & 81.5 & 80.9 & 81.4 & 78.9 & 84.3\\
 & 12B IT & \underline{95.0} & \underline{86.3} & \underline{94.2} & \underline{93.7} & \underline{93.9} & \underline{94.0} & \underline{90.8} & \underline{93.6} & \underline{94.6} & \underline{90.3} & \underline{92.3} & \underline{92.5} & \underline{92.2} & \underline{92.4} & \underline{90.3} & \underline{92.7}\\
\rowcolor{PCTTint}  & \pct{} & \textbf{96.6} & \textbf{90.3} & \textbf{95.7} & \textbf{95.4} & \textbf{95.8} & \textbf{95.8} & \textbf{92.7} & \textbf{96.9} & \textbf{96.7} & \textbf{94.5} & \textbf{94.7} & \textbf{94.6} & \textbf{95.4} & \textbf{95.3} & \textbf{93.0} & \textbf{95.9}\\
\addlinespace[2pt]
fa & 4B IT & 79.9 & 71.8 & 78.4 & 78.6 & 78.8 & 78.4 & 75.7 & 81.7 & 81.1 & 76.2 & 78.9 & 80.0 & 79.6 & 79.0 & 77.3 & 80.0\\
 & 12B IT & \underline{94.5} & \underline{86.9} & \underline{94.0} & \underline{93.7} & \underline{93.5} & \underline{93.8} & \underline{90.8} & \underline{93.1} & \underline{95.3} & \underline{91.8} & \underline{93.0} & \underline{93.6} & \underline{94.1} & \underline{93.6} & \underline{91.5} & \underline{93.7}\\
\rowcolor{PCTTint}  & \pct{} & \textbf{95.3} & \textbf{88.3} & \textbf{95.5} & \textbf{95.0} & \textbf{95.3} & \textbf{95.6} & \textbf{92.8} & \textbf{95.7} & \textbf{96.5} & \textbf{94.7} & \textbf{94.8} & \textbf{94.7} & \textbf{95.4} & \textbf{95.3} & \textbf{93.4} & \textbf{95.3}\\
\addlinespace[2pt]
es & 4B IT & 88.0 & 80.9 & 89.8 & 89.2 & 88.3 & 89.1 & 85.6 & 91.5 & 89.9 & 87.7 & 89.5 & 89.3 & 89.3 & 89.8 & 87.8 & 89.6\\
 & 12B IT & \textbf{97.7} & \underline{91.7} & \textbf{97.2} & \textbf{96.9} & \underline{96.7} & \underline{96.8} & \underline{93.7} & \underline{94.1} & \underline{95.1} & \underline{93.4} & \underline{93.4} & \underline{93.5} & \underline{93.6} & \underline{93.7} & \underline{92.0} & \underline{92.7}\\
\rowcolor{PCTTint}  & \pct{} & \underline{97.4} & \textbf{92.0} & \underline{97.0} & \underline{96.7} & \textbf{97.2} & \textbf{97.2} & \textbf{94.1} & \textbf{97.2} & \textbf{96.9} & \textbf{95.1} & \textbf{96.2} & \textbf{95.7} & \textbf{95.6} & \textbf{96.7} & \textbf{94.3} & \textbf{95.5}\\
\addlinespace[2pt]
\bottomrule\end{tabular}
\caption{All eight judge dimensions in both views, aggregated by target language, with 50 shared examples per target. All directions receive equal weight.}
\end{table*}
\clearpage
\paragraph{Metric settings.} SacreBLEU 2.5.1: one reference, mixed-case BLEU, exponential smoothing, no effective order, \texttt{zh} tokenization for Chinese targets and \texttt{13a} otherwise. chrF and chrF++ use character order six, $\beta=2$, and respectively zero/two word orders, excluding character whitespace. TER uses lowercase Tercom tokenization with normalization, punctuation retained, and Asian support. The compiled TER beam kernel preserves the reference operation tie order, beam, shift limits, and prefix cache; exact edit/length totals match SacreBLEU on all 900 judge-subset outputs, alongside 2,000 exact score/trace cache checks and 250 candidate-order checks. The compiled complete search additionally matches 500 randomized cases; the reference tokenizer and search semantics are retained.
\paragraph{COMET.} We use \texttt{Unbabel/wmt22-comet-da} with unbabel-comet 2.2.7, official checkpoint revision \texttt{2760a223ac95}, and verified SHA-256. Inference uses float32 (batch 32 for full-test scoring, 16 for the judge subset). Values in tables are raw scores multiplied by 100 and are not accuracy percentages. The encoder truncates each source, candidate, and reference separately at 512 tokens, so long-document suffixes are not assessed by this metric.
\begin{table}[t]\centering\small
\setlength{\tabcolsep}{2.8pt}
\renewcommand{\arraystretch}{1.08}
\begin{tabular}{llrrrr}
\toprule
\rowcolor{TableHeaderDark} {\color{white}\bfseries Split} & {\color{white}\bfseries System} & {\color{white}\bfseries $N$} & {\color{white}\bfseries Src.} & {\color{white}\bfseries Cand.} & {\color{white}\bfseries Ref.}\\
\midrule
Full & 4B IT & 36,572 & 7757 & 8240 & 7789\\
Full & 12B IT & 36,572 & 7757 & 8334 & 7789\\
\rowcolor{PCTTint} Full & \pct{} & 36,572 & 7757 & 7743 & 7789\\
Judge & 4B IT & 300 & 51 & 52 & 52\\
Judge & 12B IT & 300 & 51 & 53 & 52\\
\rowcolor{PCTTint} Judge & \pct{} & 300 & 51 & 51 & 52\\
\bottomrule\end{tabular}
\caption{Texts exceeding the 512-token COMET encoder limit. Src., Cand., and Ref. denote source, candidate, and reference. Counts can overlap within an example.}
\end{table}